\documentclass{article}

\usepackage{arxiv}

\usepackage{amsmath}

\usepackage{url,lineno,microtype}
\usepackage[onehalfspacing]{setspace}
\usepackage{tikz}
\usetikzlibrary{shapes.geometric, arrows}
\usepackage{bbding}
\usepackage{multirow}
\usepackage{tabularx,ragged2e}
\usepackage{booktabs}
\usepackage{amssymb}
\usepackage{float}
\usepackage{subfig}
\usepackage{graphicx}
\usepackage{anyfontsize}
\usepackage{bm}
\usepackage{eurosym}
\usepackage{diagbox}
\usepackage{array}
\usepackage{lscape}
\usepackage{caption}
\usepackage{xcolor, colortbl}
\usepackage{lscape}
\usepackage{array, multirow, multicol}
\usepackage[hidelinks]{hyperref}
\tikzstyle{startstop} = [rectangle, rounded corners, minimum width=3cm, minimum height=1cm, text centered, draw=black, fill=red!30]
\tikzstyle{process} = [rectangle, minimum width=3cm, minimum height=1cm, text centered, draw=black, fill=orange!30]
\tikzstyle{decision} = [diamond, minimum width=3cm, minimum height=1cm, text centered, draw=black, fill=green!30]
\tikzstyle{arrow} = [thick,->,>=stealth]

\newcolumntype{b}{>{\hsize=.5\hsize}X}
\newcolumntype{s}{>{\hsize=.4\hsize}X}
\newcolumntype{C}[1]{>{\centering\let\newline\\\arraybackslash\hspace{0pt}}m{#1}}
\title{Swine Diet Design using Multi-objective Regionalized Bayesian Optimization} 

\author{ \href{https://orcid.org/0009-0001-2411-5000}{\includegraphics[scale=0.06]{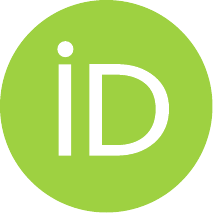}\hspace{1mm}Gabriel D. Uribe-Guerra}\\
	Intelligent Information Systems Lab\\
	Department of Systems Engineering\\
        Universidad de Antioquia\\
	Calle 67 No. 53 - 108, Medellín, Colombia. \\
	\texttt{gdario.uribe@udea.edu.co} \\
	\And
	\href{https://orcid.org/0000-0003-0762-0571}{\includegraphics[scale=0.06]{orcid.pdf}\hspace{1mm}Danny A. Múnera-Ramírez} \\
	Intelligent Information Systems Lab\\
	Department of Systems Engineering\\
        Universidad de Antioquia\\
	Calle 67 No. 53 - 108, Medellín, Colombia. \\
	\texttt{danny.munera@udea.edu.co} \\
        \And
        \href{https://orcid.org/0000-0002-1928-773X}{\includegraphics[scale=0.06]{orcid.pdf}\hspace{1mm}Julián D. Arias-Londoño}\thanks{Corresponding author.}\\
	Department of Signals, Systems, and Radiocommunications\\
	ETSI Telecomunicación\\
        Universidad Politécnica de Madrid\\
	Av. Complutense, 30, 28040, Madrid, Spain. \\
	\texttt{julian.arias@upm.es} \\
}

\renewcommand{\shorttitle}{Swine Diet Design using Multi-objective Regionalized Bayesian Optimization}

\hypersetup{
pdftitle={A template for the arxiv style},
pdfsubject={q-bio.NC, q-bio.QM},
pdfauthor={David S.~Hippocampus, Elias D.~Striatum},
pdfkeywords={First keyword, Second keyword, More},
}

\begin{document}
\maketitle
\begin{abstract}
The design of food diets in the context of animal nutrition is a complex problem that aims to develop cost-effective formulations while balancing minimum nutritional content. Traditional approaches based on theoretical models of metabolic responses and concentrations of digestible energy in raw materials face limitations in incorporating zootechnical or environmental variables affecting the performance of animals and including multiple objectives aligned with sustainable development policies. Recently, multi-objective Bayesian optimization has been proposed as a promising heuristic alternative able to deal with the combination of multiple sources of information, multiple and diverse objectives, and with an intrinsic capacity to deal with uncertainty in the measurements that could be related to variability in the nutritional content of raw materials. However, Bayesian optimization encounters difficulties in high-dimensional search spaces, leading to exploration predominantly at the boundaries. This work analyses a strategy to split the search space into regions that provide local candidates termed multi-objective regionalized Bayesian optimization as an alternative to improve the quality of the Pareto set and Pareto front approximation provided by BO in the context of swine diet design. Results indicate that this regionalized approach produces more diverse non-dominated solutions compared to the standard multi-objective Bayesian optimization. Besides, the regionalized strategy was four times more effective in finding solutions that outperform those identified by a stochastic programming approach referenced in the literature. Experiments using batches of query candidate solutions per iteration show that the optimization process can also be accelerated without compromising the quality of the Pareto set approximation during the initial, most critical phase of optimization.
\end{abstract}


\section{Introduction}

Food diet design for livestock and poultry production is an increasingly complex problem that, in principle, aims to develop diet formulations that can balance minimum nutritional contents with cost-effective food production criteria \cite{barszcz2024poultry}. However, farmers progressively have had to wrestle with additional obstacles, such as preserving production efficiency in constantly changing environmental contexts resulting from climate change effects and persistent growth in global food demand \cite{herrero2023livestock, akintan2024animal}. Climate change imposes new challenges to the livestock industry because it can affect raw material and agricultural by-products' availability, affecting soil salinity and reducing forage areas, but also increasing heat stress of the animals, which limits the capacity of livestock to shed heat to their environment \cite{godde2021impacts}, and results in reduced feed intake and body weight gain \cite{rauw2017effects}. Simultaneously, livestock farming stands as a significant contributor to greenhouse gas emissions, so there's a growing push for the industry to develop dietary strategies aimed at shrinking its ecological impact~\cite{mackenzie2016towards}.

Considering the multiple objectives and restrictions that must be satisfied simultaneously, during the last decades, the food diet design problem has been addressed as a single/multi-objective constrained optimization problem \cite{uyeh2019precision}. Techniques employed includes classical Linear Programming (LP) \cite{chappell1974linear,glen1980mathematical,sebastian2008simple}, Stochastic Programming (SP) \cite{patil2022stochastic,pena2009multiobjective}, Non-Linear Programming (NLP) \cite{afrouziyeh2010use, jardim2013application} and different heuristic approaches \cite{gupta2013heuristic,csahman2009cost, altun2013cost}. Moreover, the considered objectives include productivity, weight or conversion maximization, reduction of excretions, energy density, variations in nutrient reduction, and greenhouse gas cutback, among others \cite{ghosh2014current, d1992least}; some of them have been addressed at the same time. 

The evolution of the applied techniques and focus over time is attributed to various technical limitations, such as the inability to apply solutions across different breeds or species universally and the need to deal with the variability in raw materials or animal metabolism \cite{uyeh2019precision}. The initial designs prioritized meeting constraints at the lowest cost, regardless of the species involved. In \cite{chappell1974linear}, the focus was on creating a practical system to help ingredient buyers effectively use LP in real-world scenarios, thus advancing diet design systems. Similarly, \cite{glen1980mathematical} applied LP to improve operational efficiency and decision-making processes in the cattle industry. Additionally, \cite{sebastian2008simple} utilized LP to minimize costs on small dairy farms by using low-cost raw materials that meet nutritional requirements.

However, the challenges of achieving efficient production and reducing nutrient variability to ensure environmental friendliness in the industry have led to LP being set aside in favor of exploring new techniques. Consequently, SP techniques have been proposed as a viable alternative, ensuring optimal nutrient levels in formulations at minimal cost, especially for lactating cattle. In \cite{patil2022stochastic}, authors demonstrated that using SP in diet formulation reduces variability and provides a confidence level exceeding $90\%$ for meeting nutrient requirements. Similarly, \cite{pena2009multiobjective} suggested SP to address a multi-objective problem in pig nutrition, focusing on assuring specific amounts of lysine and metabolizable energy while minimizing nutrient variability at the lowest cost.

In addition to reducing nutrient variability in formulation ingredients, efforts have been made to enhance productivity, leading to the use of NLP models. In \cite{afrouziyeh2010use}, this technique was applied to determine optimal nutrient concentrations for maximizing egg bird productivity while minimizing costs. NLP  techniques were also used to evaluate the intake and digestibility of food in sheep at different stages of nutrition to improve food quality \cite{jardim2013application}.

In recent years, heuristic approximation methods have enabled the inclusion of variables beyond dietary and nutritional factors, allowing for the evaluation of additional inputs that influence the outcomes of formulation systems. Studies, such as \cite{csahman2009cost}, illustrate that Genetic Algorithms (GA) for optimizing poultry diet formulation costs are more efficient than LP algorithms. Similarly, \cite{gupta2013heuristic} used GAs to efficiently optimize formulations for dairy cows by adjusting nutritional objectives. This work also shows that GA helps assess consequences such as rising input prices and environmental impact. Additionally, \cite{altun2013cost} utilized Particle Swarm Optimization (PSO) and Real-Coded GA (RCGA) to study nonlinear feed mix problems for poultry and other animals like rabbits, goats, and cows. PSO showed faster and more effective performance in solving nonlinear feed formulation problems compared to RCGA. Although GA and PSO are flexible enough to perform the optimization of complex non-analytic non-linear problems and to combine variables from different sources, their main drawback for addressing the diet design problem is the limited optimization budget in real experimental-sampling approaches due to the highly expensive evaluation of the optimization function. Therefore, in \cite{guribe2024}, the authors explore the use of Bayesian Optimization (BO) to reduce variability in the quality of diet formulation while reducing costs and ensuring nutrient content. Their results demonstrated promising results in the quality of the solutions without requiring a large number of function evaluations, in addition to the inherent advantage of BO in dealing with the variability in nutrient content through data noise estimates. 

Notwithstanding its advantages, solving food diet design problems using BO faces additional challenges \cite{guribe2024}. Apart from the requirements of incorporating contextual (non-controllable) variables~\cite{krause2011contextual} and the need for dealing with data coming from different farms and fidelity levels \cite{belakaria2021output}, the size of the search spaces, which includes zootechnic, environmental and ingredient-related variables, exceeds the dimensions for which standard BO techniques expose a good balance between exploration and exploitation traits \cite{eriksson2019scalable}. The optimization of high-dimensional problems is complicated because since the search spaces grow exponentially with the dimension, they could become more plentiful of local optima, global optima are more challenging to find, and, due to the curse of dimensionality and limited sampling budgets, regions with large posterior uncertainty proliferate, making the BO algorithm to emphasize less in the exploitation of promising areas \cite{eriksson2019scalable}. Moreover, in high-dimensional spaces, such a preference of BO for exploring concentrates on the boundaries of the search space \cite{wang2020learning}, which makes it even more complex to find global optima.

To overcome this problem within the domain of BO, three primary methodologies have been suggested: i) using a low-dimensional transformation of the feature space, including linear \cite{zhang2019high} and non-linear \cite{oh2018bock} mappings or embeddings; or a simultaneous variable selection and optimization to learn the underlying effective subspace \cite{chen2012joint}; ii) by establishing an additive structure of multiple lower-dimensional subproblems and incorporating a message passing phase during the optimization process \cite{rolland2018high,han2021high,ziomek2023random}; iii) implementing a space partitioning strategy based on a trust region (TR) BO algorithm \cite{eriksson2019scalable}, wherein the search space is split into different subregions and local models are fitted to each TR; or by applying a meta-level algorithm termed LA-MCTS that recursively learns space partition hierarchically \cite{wang2020learning}. Apart from inherent computational complexity drawbacks, the first two approaches may encounter challenges when the input space combines discrete and continuous variables, a scenario common in zootechnical variables like inheritance traits or growth phase. Moreover, they may lack proper correlation patterns among variables due to the specific decomposition applied to the target function. Regarding the third group of approaches, while LA-MCTS offers a strategy to partition the search space more informatively, it combines a large set of supervised and unsupervised algorithms to learn a binary tree (a $k$-means clustering and a Support Vector Machine\textemdash SVM per node), whose leaves are refined using the same algorithm proposed in \cite{eriksson2019scalable}, making the whole approach relies heavily on the selection of several hyperparameters and computationally expensive. Additionally, the SVM-based partition strategy does not have a straightforward generalization to multi-objective scenarios. In contrast, the TR-based BO optimization algorithm, called TURBO, employs a straightforward strategy of partitioning the search space into different hyperrectangles centered at the best solution found so far. It trains a surrogate model per region and uses a Thomson sampling strategy to select the best candidate from realizations of the posterior functions from each local GP, thereby making the exploitation-exploration trade-off more efficient without affecting or transforming the input space. This approach was extended to multi-objective problems in \cite{daulton2022multi}, where authors proposed the Multi-Objective Regionalized Bayesian Optimization (MORBO) algorithm, which uses the same principles of TURBO to optimize diverse parts of the global Pareto frontier in parallel using a coordinated set of local TR. Additionally, unlike TURBO, MORBO implements an information-sharing policy among the TRs to facilitate efficient and collaborative global optimization \cite{daulton2022multi}. 

Bearing this in mind, this paper aims to demonstrate empirically the performance improvement resulting from applying a 
MORBO approach to an existing multi-objective swine diet design problem, comprising 17 input variables related to ingredient proportions in the formulation, 28 zootechnical and nutritional constraints, and three objectives: energy content, lysine, and cost. This is a step forward in the adaptability and suitability evaluation of BO-based methods to tackle diet design problems in real scenarios where the cost function sampling involves evaluating a specific diet design for feeding a pig herd, with variables from various sources combined to model pig performance. Similar to the analysis carried out in our previous work \cite{guribe2024}, this paper analyses the most critical hyperparameters of the MORBO strategy, performs a systematic analysis of the Thomson sampling strategy for selecting the best candidates and the effects of the number and size of the TRs. The Pareto front approximation is evaluated in terms of hypervolume, cardinality, and diversity. Also, an analysis of the exploration performed by MORBO vs. that from the standard MOBO method is presented regarding the three spaces involved: input space (ingredients), nutrient space, and output space (target). Moreover, a low-dimensional projection strategy is also implemented to compare the diversity observed in the Pareto set of MORBO and MOBO. A comparative evaluation of the quality of solutions provided by MORBO regarding those obtained using MOBO and those proposed in \cite{pena2009multiobjective} is included. Results of speeding up the optimization process by querying the algorithm to private a batch of solutions are also presented.

The rest of the paper is organized as follows: Section \ref{sec:problem} outlines the mathematical problem and provides a detailed explanation of the solution proposed in \cite{pena2009multiobjective}. Section \ref{sec:methods} introduces BO, the multi-objective variant and a regionalized approach for high-dimensional problems; Section \ref{sec:exper_res} details the experiments conducted, the performance measures used, and the outcomes achieved. Finally, Section \ref{sec:concl} presents the conclusions drawn from the research.

%The rest of the paper is organized as follows: section \ref{sec:problem} presents the mathematical formulation of the objective problem and describes in more detail the solution proposed in \cite{pena2009multiobjective}; section \ref{sec:methods} introduces BO, the Multi-objective variant and the regionalized approach for high-dimensional problems;  section \ref{sec:exper_res} describes the set of experiments, the evaluation metrics and presents the results obtained. Lastly, section \ref{sec:concl} provides some conclusions from the results.

\section{Problem Statement}\label{sec:problem}

% 1 model: MO lys energi costo ->  lineal -> variabilidad 
% 2 métodos peña

The swine diet design problem addressed in this work was introduced in \cite{pena2009multiobjective}, where the authors proposed a multi-objective optimization challenge centered on nourishment formulations considering two crucial factors: the nutritional value and the cost of feed composition. The aim is to ensure the presence of essential nutrients in pig diets while maintaining economic viability. Lysine and energy emerged as pivotal elements for achieving nutritional balance in these diets.

Lysine, an essential amino acid, is crucial for nutrient metabolism and protein synthesis, and vital for animal growth. Meanwhile, energy is fundamental to life processes. Research indicates that an imbalance in these nutrients can impair pig development \cite{cho2012effect}.
%Studies showing that an imbalance in these nutrients can hinder pig performance \cite{cho2012effect} underscores the intricate relationship between lysine and energy.

Consequently, the optimization challenge involves identifying the optimal blend and proportions of dietary components. The goal is to maximize lysine and energy content while minimizing costs. This necessitates imposing constraints on essential nutrients such as crude fiber, phosphorus, calcium, and protein, as outlined in \cite{pena2009multiobjective}.

%Consequently, the optimization problem seeks to determine the best combination of ingredients and their proportions within the diet. The goal is to maximize the availability of lysine and energy while minimizing costs. This endeavor entails imposing various constraints on nutritional requirements, including but not limited to crude fiber, phosphorus, calcium, and crude protein, as detailed in \cite{pena2009multiobjective}.

The ingredients, nutritional values, and costs were determined according to the guidelines set by the Spanish Foundation for the Development of Animal Nutrition (FEDNA) in 1999, as outlined in \cite{pena2009multiobjective}. These specifications were utilized to ensure consistency, and for comparative purposes, the same tables of nutrient content and costs were adopted for this study.

Formally, the problem can be formulated as \cite{guribe2024}:

\begin{equation}\label{eq:opt_problem}
\begin{aligned}
\text{min} \quad & f_c({\bf{x}}) = {\bf{c}}^T {\bf{x}} \\
\text{max} \quad & f_l({\bf{x}}) = {\bf{l}}^T {\bf{x}} \\
\text{max} \quad & f_e({\bf{x}}) = {\bf{e}}^T {\bf{x}} \\
\text{s.t} \quad & {\bf 1}^T {\bf{x}} =1 \\
&  \underline{\bf{b}}\leq {\bf{A}}^T{\bf{x}}\leq \overline{\bf{b}}\\
& 0\leq  {\bf{x}} \leq  {\bf{s}}\\
\end{aligned}
\end{equation}

where ${\bf{x}}=[x_1,x_2,\cdots,x_i,\cdots,x_{d}]$ represents a vector solution containing the proportions of each ingredient in a diet comprising $d$ ingredients. ${\bf{A}}$ is a matrix describing the nutrient composition of each ingredient with respect to a set of $a$ nutrients, while ${\bf{c}}$ is a $d$-dimensional vector indicating the costs associated with each ingredient. Vectors ${\bf{l}}$ and ${\bf{e}}$ encompass the lysine and energy content of each ingredient, respectively. Furthermore, $\underline{\bf{b}}$ and $\overline{\bf{b}}$ denote vectors stipulating the lower and upper bounds for each nutrient, and ${\bf{s}}$ signifies a vector representing the upper bounds for each raw material during diet formulation. As previously stated, ${\bf{A}}$, ${\bf{l}}$, and ${\bf{c}}$ conform to FEDNA tables \cite{pena2009multiobjective}.

Based on the original formulation by Noblet \cite{noblet2004energy}, the model for digestible energy $f_e$ employs an equation linking the diet's nutritional composition with energy, thus defining the vector ${\bf{e}}$. The comprehensive inventory of ingredients and nutrients, as detailed in \cite{pena2009multiobjective}, can be found in~\ref{sec:appe1}.

Peña and colleagues proposed a semi automatic method called Mult-Objetive Fractional problem (MFP) to address the optimization as mentioned earlier formulation \cite{pena2009multiobjective}. This method aims to create economically viable animal diets while reducing variability in the nutrients content of ingredients. The MFP takes into account both the cost of rations and the probability of meeting the animal's nutritional needs. To solve the MFP, the authors introduced the Interactive Multi-Objective Goal Programming (IMGP) method \cite{spronk1981interactive}, which involves a decision-maker who expresses preferences without needing prior knowledge of the minimum probabilities required to meet nutritional requirements.
%This approach focuses on crafting economically feasible animal diets while mitigating variability in nutrient content among raw materials. The MFP integrates considerations of both ration costs and the likelihood of meeting the animal's nutrient needs. To solve the MFP, the authors introduced the Interactive Multi-Objective Goal Programming (IMGP) method  \cite{spronk1981interactive}, which involves a decision-maker who articulates preferences without requiring prior knowledge of the minimum probabilities needed to fulfill nutritional requirements.

In the IMPG approach, each of the three objectives is optimized independently without considering the constraints of the other objectives, to establish the best and worst values. These values define the interval for establishing objective goals. The decision maker selects an objective to improve and all objectives are optimized again, incorporating the restrictions of the objectives with the established limits. The decision maker evaluates whether the improvements in the chosen objective justify possible adjustments in the others, making the corresponding modifications. In this process, the constraints evolve in each iteration, gradually reducing the range of feasible solutions, until the lower and upper limits of the objectives converge. This convergence means an efficient solution in which improving one objective no longer detracts from another.
%In the IMGP approach, the three objectives are initially optimized independently, establishing optimal and least favorable values for each. These values then delineate the range for defining objective goals in subsequent iterations. The decision-maker selects an objective for enhancement, and the objectives undergo optimization again, incorporating constraints from the original problem and new constraints based on the defined limits. The decision-maker evaluates whether improvements in the chosen objective warrant potential adjustments in others and modifications are made accordingly. Constraints evolve with each iteration, progressively narrowing the scope of feasible solutions until the lower and upper bounds of objectives converge. This convergence indicates an efficient solution where no alternative enhances one objective without deteriorating another.

\section{Methods}\label{sec:methods}

\begin{figure*}[ht]
    \hspace{-2cm}
    \includegraphics[scale=0.6]{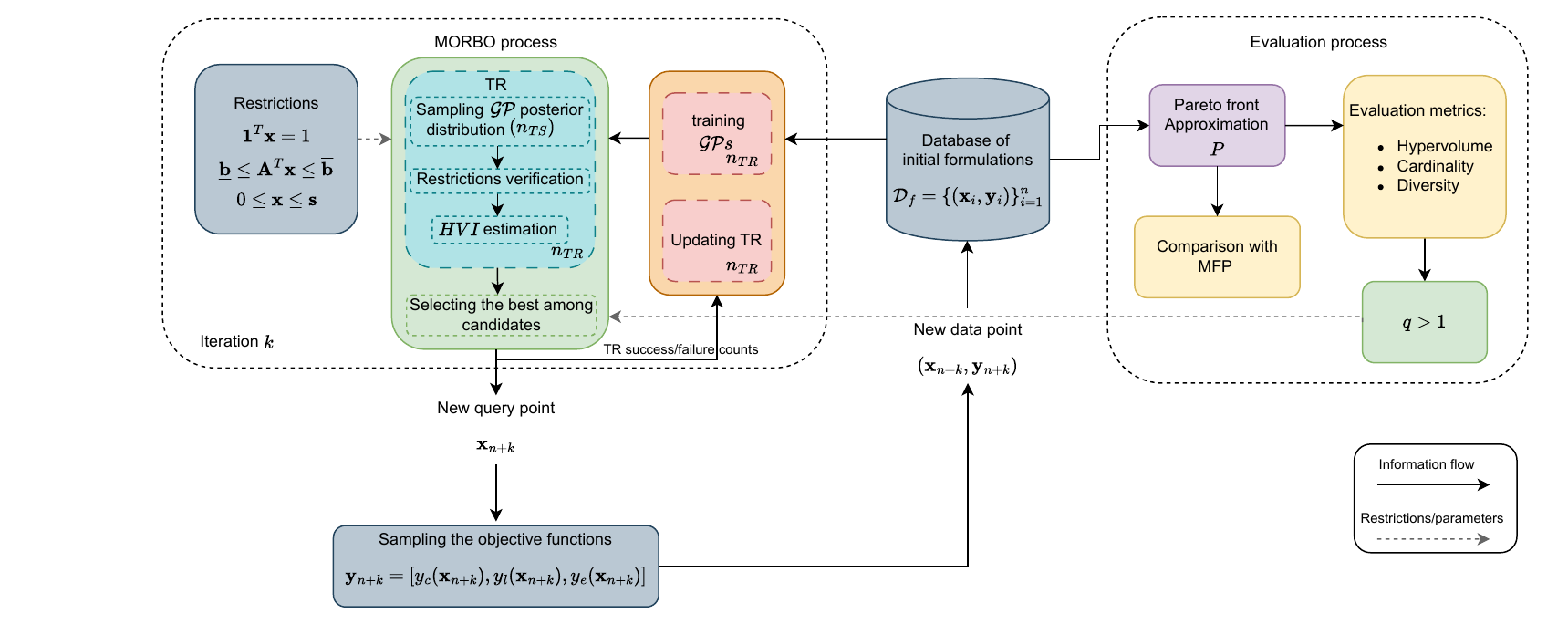}
    \caption{General scheme of the proposed methodology. Image adapted from \cite{guribe2024}.}
    \label{fig:scheme}
\end{figure*}

In this section, we outline the methods and techniques used to optimize the swine diet formulation problem discussed earlier, employing both the standard Multi-Objective Bayesian Optimization (MOBO) and the Multi-Objective Regionalized Bayesian Optimization (MORBO) approaches. Figure \ref{fig:scheme} provides a schematic representation of the proposed methodology based on MORBO, highlighting each component and its function throughout the process.
%In this section, we present the methods and techniques employed to optimize the swine diet formulation problem discussed earlier, utilizing both the standard Multi-Objective Bayesian Optimization (MOBO) and Multi-Objective Regionalized Bayesian Optimization (MORBO) approaches. Figure \ref{fig:scheme} presents a schematic representation of the proposed methodology based on MORBO, identifying each component and its role throughout the process.

\subsection{Bayesian Optimization}

BO is a machine learning-driven optimization technique, intended for cases where the continuous objective function $f({\bf{x}})$ lacks an analytical form or is expensive to evaluate, and where computing the derivatives of first and second-order is not feasible \cite{frazier2018tutorial}. BO has two main components: a Bayesian surrogate model for the objective function and an acquisition function (AF) to identify the next sampling location. The fundamental idea of BO methods is to use the posterior distribution of the Bayesian model to explore the search space $\mathcal{X} \subset \mathbb{R}^d$ and choose input values $\bf{x}$ that Sean probably maximizes the objective function $f$ in the objective space~$\mathcal{Y}$.
%BO is a machine learning-driven optimization technique created for situations where the continuous objective function $f({\bf{x}})$ is costly to assess, lacks an analytic expression, and where obtaining first and second-order derivatives of $f$ is impractical \cite{frazier2018tutorial}. BO comprises two main elements: a Bayesian surrogate model to represent the objective function and an acquisition function (AF) to determine the next sampling location. The fundamental concept underlying BO methods is to leverage the posterior distribution of the Bayesian model to explore the search space $\mathcal{X} \subset \mathbb{R}^d$ and choose input values $\bf{x}$ likely to maximize the objective function $f$ in the target space~$\mathcal{Y}$. 

In BO, Gaussian Processes (GPs) are frequently used as surrogate models because they provide a principled way to evaluate uncertainty and have a flexible \cite{williams2006gaussian} core design. There are also other alternatives available \cite{listudy2024}. The process begins with an initial set of observations $\mathcal{D}{f}=\lbrace \left({\bf x}{1}, y_{1}\right), \left({\bf x}{ 2}, y{2}\right), \dots, \left({\bf x}{n}, y{n}\right) \rbrace$, which are assumed to be corrupted with additive Gaussian noise, $y_i = f({\bf x}_{i}) + \varepsilon$, where $\varepsilon \sim \mathcal{N}(0, \sigma^2)$.

%In BO, surrogate models commonly rely on Gaussian Processes (GPs) due to their capacity for providing a principled and manageable assessment of uncertainty alongside their adaptable kernel design \cite{williams2006gaussian}, although other alternatives exist \cite{listudy2024}. The process starts with an initial set of observations $\mathcal{D}_{f}=\lbrace \left({\bf x}_{1},y_ {1}\right),\left({\bf x}_{2},y_{2}\right), \dots, \left({\bf x}_{ n},y_{n}\right) \rbrace$, assumed to be corrupted with additive Gaussian noise, $y_i = f({\bf x}_{i}) + \varepsilon$, where $\varepsilon \sim \mathcal{N}(0,\sigma^2)$ must be available.

From the initial dataset $\mathcal{D}_{f}$, the standard Bayesian optimization process comprises three steps:

\begin{enumerate}
    \item Construct a GP model using $\mathcal{D}_{f}$.
    \item Utilize the posterior distribution over $f$ provided by the trained GP to guide the AF, which determines areas in the search space worthy of exploitation or exploration. Consequently, regions where $f({\bf{x}})$ is optimal or unexplored areas with the potential to enhance the current best solution receive a high AF value.
    
    \item Apply an optimization algorithm to the AF to identify the next query point ${\bf x}_{n+1}$. Once sampled from the target function $f({\bf x}_{n+1})$, it can be appended to the history of observations $\mathcal{D}_{f}=\mathcal{D}_{ f} \cup ({\bf x}_{n+1}, y_{n+1})$.

\end{enumerate}

This process iterates until the optimization budget is depleted. The size $n$ of the initial set, which corresponds to previously known diet formulations, represents a critical hyperparameter in this process. In the following sections, we describe the essential components of this optimization process.

\subsubsection{Gaussian processes}

A GP consists of a set of random variables that, for certain finite subsets, follow a joint Gaussian distribution \cite{williams2006gaussian}. Consequently, a GP characterizes a distribution over functions $f(\cdot)\sim \mathcal{GP}(\mu(\cdot), k(\cdot,\cdot))$, where parameters are assigned to the mean $\mu(\cdot)$ and the kernel $k(\cdot, \cdot)$. This distribution is defined for each pair of points  ${\bf x, x'} \in \mathbf{R}^d$ and represents the covariance between them. Thus:
\begin{align*}
    \mu({\bf x})&=\mathbb{E}\left[ f({\bf x})\right]\\
    k({\bf x},{\bf x'})&=\mathbb{E}\left[(f({\bf x})-\mu({\bf x}))(f({\bf x'})-\mu({\bf x'})\right)]
\end{align*}
Standard practice includes adjusting the kernel function's hyperparameters by maximizing the marginal likelihood across the training set, with full Bayesian methods being another viable option \cite{lalchand2020approximate}.

Using a set of input samples $X = \lbrace{{{\bf x}{1}}, \dots, {{\bf x}{n}}\rbrace}$ and their corresponding noisy outputs $Y = \lbrace{y_{1}, \dots, y_{n}\rbrace}$, the posterior distribution of a new point $\hat{{\bf x}}$ can be obtained from the joint Gaussian distribution:

$$\begin{bmatrix}
Y\\ \hat{y} 

\end{bmatrix}=\mathcal{N}\left( \begin{bmatrix}
\mu(X)\\ \mu(\hat{{\bf x}}) 

\end{bmatrix}, \begin{bmatrix}
k(X,X)+ \sigma^{2}I & k( X,\hat{{\bf x}})\\ 
k(\hat{{\bf x}},X) & k(\hat{{\bf x}},\hat{{\bf x}})
\end{bmatrix}\right)$$

Subsequently, the posterior distribution of $f(\cdot)$ at the point ${\bf \hat{x}}$, denoted as $f({\bf \hat{x}})=p(y|{\bf \hat{x}},X,Y)$, can be estimated utilizing the conventional conditioning rules for Gaussian random variables, yielding $f({\bf \hat{x}})\sim \mathcal{N}\left( \mu_{f}({\bf \hat{x}}), \sigma_{f}({\bf \hat{x}})\right)$, where

\begin{align*}
    \mu_{f}({\bf \hat{x}})&=\mu(\hat{{\bf x}})+k(\hat{{\bf x}},X)\left[k(X,X)+\sigma^{2}I\right]^{-1}(Y-\mu(X))\\
    \sigma_{f}({\bf \hat{x}})&=k(\hat{{\bf x}},\hat{{\bf x}})-k(\hat{{\bf x}},X)\left[k(X,X)+\sigma^{2}I\right]^{-1} k(X,\hat{{\bf x}})
\end{align*}

For a more comprehensive understanding of GPs, please refer to \cite{williams2006gaussian}.

\subsubsection{Acquisition functions}

Acquisition functions exploit the posterior mean and variance at each function point to compute a metric that indicates how beneficial it would be to sample at that point in the next iteration. The goal of an effective acquisition function is to balance exploration and exploitation.

Several acquisition functions are commonly used in the literature, including the Upper Confidence Bound (UCB) \cite{srinivas2009gaussian}, the Probability of Improvement (PI) \cite{kushner1964new}, and the Expected Improvement (EI) \cite{movckus1975bayesian}. UCB balances exploration and exploitation by combining the mean and variance of the posterior distribution. PI selects points to maximize the probability of achieving a better objective value than the current best. EI, the most widely employed acquisition function, evaluates points based on both the probability and the magnitude of improvement, aiming to find the point that provides the greatest expected enhancement to $f$.

%Numerous AFs have been proposed in the literature, with some of the most commonly used including the upper confidence bound (UCB) \cite{srinivas2009gaussian}, the Probability of Improvement (PI) \cite{kushner1964new}, and the Expected Improvement (EI) \cite{movckus1975bayesian}. UCB achieves a balance between exploitation and exploration by considering the sum of the mean and variance of the posterior distribution. In contrast, PI selects the next query sample to maximize the probability of achieving an objective value higher than the current optimum. The most widely used AF is EI, which not only accounts for the probability of improvement but also the magnitude of improvement. It evaluates $f$ at the point that, on average, yields the greatest enhancement to $f$. 

While the standard BO framework provides one new query point per iteration, there are advanced methods that extend BO to provide multiple candidate solutions for accelerated optimization. For instance, parallel EI ($q$-EI) evaluates improvement over a joint probability distribution of $q$ points and selects the set of points that maximizes the expected improvement across these points \cite{ginsbourger2010kriging}.

\subsection{Multi-objective Bayesian Optimization}

BO can be seamlessly extended to vector-valued functions by defining and optimizing a performance metric over sets, using EI to guide the search process across multiple objectives. Nonetheless, this adds complexity to the problem due to the numerous directions in which objectives can be improved \cite{shu2020new}. In any multi-objective strategy, the main aim of MOBO is to find a set of points that represent the best trade-offs between $m$ conflicting objectives, commonly referred to as the Pareto set $X_p = {{\bf x}_1, {\bf x}_2, \cdots, {\bf x}_t}, ; {\bf x}_i \in \mathcal{X}$, along with the corresponding Pareto front ($\mathcal{P}$), which is a set of solutions $\mathcal{P} = {{\bf y}_1,{\bf y}2,\cdots,{\bf y}t}$ in the objective space, where ${\bf y}j = [y{j1},y{j2},...,y{jm}]^T = {\bf f}({\bf x}_j) + {\bm{\varepsilon}}$, and $\nexists ; {\bf y}_l | {\bf y}_l \succ {\bf y}_j, l \neq j, ; \forall {\bf y}_l \in \mathcal{Y} \subset \mathbb{R}^m$ \cite{galuzio2020mobopt}. Here, the symbol $\succ$ signifies dominance, meaning that ${\bf y}_l$'s objective values are at least as good as those of ${\bf y}j$, and better in at least one objective. In MOBO, $\mathcal{P}$ is approximated by a set $P$ containing all non-dominated solutions in $\mathcal{D}{\bf f}$.

%Using EI, BO can be straightforwardly extended to vector-valued functions by defining and improving a performance metric over sets, which can be used to guide the search process across multiple objectives. However, this increases the problem's difficulty because there are many directions in which the objectives can be improved \cite{shu2020new}. As in any multi-objective approach, the ultimate goal of MOBO is to identify a collection of points that describes the best trade-offs among $m$ different conflicting objectives, which is typically called the Pareto set $X_p = \{{\bf x}_1, {\bf x}_2, \cdots, {\bf x}_t\}, \; {\bf x}_i \in \mathcal{X}$, along with its corresponding Pareto front ($\mathcal{P}$), which is the set of solutions $\mathcal{P} = \{{\bf y}_1,{\bf y}_2,\cdots,{\bf y}_t\}$ in the objective space, for ${\bf y}_j = [y_{j1},y_{j2},...,y_{jm}]^T = {\bf f}({\bf x}_j) + {\bm{\varepsilon}}$, such that $\nexists \; {\bf y}_l | {\bf y}_l \succ {\bf y}_j, l \neq j, \; \forall {\bf y}_l \in \mathcal{Y} \subset \mathbb{R}^m$ \cite{galuzio2020mobopt}. The symbol $\succ$ represents dominance and, for the previous definition, would imply that objective values of ${\bf y}_l$ are no worse than those of ${\bf y}_j$, and objective values of ${\bf y}_l$ are strictly better than at least one of those of ${\bf y}_j$. In MOBO, $\mathcal{P}$ is approximated by a set $P$ of all the non-dominated solutions in $\mathcal{D}_{\bf f}$.

To maintain simplicity, this work models each objective function with independent GP priors. Each solution vector ${\bf y}_i$ represents the noisy evaluation of the three objective functions, expressed as ${\bf y}_i = [y_e({\bf x}_i), y_l({\bf x}_i), y_c({\bf x}_i)]$, where $y_e({\bf x}_i) = f_e({\bf x}_i) + \varepsilon_e$, with analogous expressions for $y_l$ and $y_c$.

%For the sake of simplicity, in this work, every objective function is modeled by independent GP priors. Each vector of solutions ${\bf y}_i$ corresponds to the noisy sampling of each of the three objective functions ${\bf y}_i = [y_e({\bf x}_i), y_l({\bf x}_i), y_c({\bf x}_i)]$, where $y_e({\bf x}_i) = f_e({\bf x}_i) + \varepsilon_e$ and similarly for $y_l$ and $y_c$.  

\subsubsection{Acquisition Functions for MOBO}\label{sec:AF_MOBO}

The most commonly used AF in MOBO seeks to estimate the expected improvement in the area under $\mathcal{P}$ generated by a new point ${\bf x}$ and its corresponding posterior distribution, which is approximated using the hypervolume (HV) indicator. First introduced in \cite{zitzler1999multiobjective}, HV is one of the key unary indicators for evaluating the quality of a Pareto front approximation set. A significant advantage of this indicator is that it does not require prior knowledge of the Pareto front. Maximizing HV can lead to a highly effective and diverse Pareto front approximation set \cite{yang2019efficient}. The HV indicator measures the volume of the subspace dominated by $\mathcal{P}$, bounded below by a reference point ${\bf r}$, and defined as:

%The most used AF in MOBO aims to estimate the expected improvement of the area under $\mathcal{P}$ given by a new point ${\bf x}$ and its corresponding posterior distribution, which is estimated based on the hypervolume (HV) indicator. HV was introduced in \cite{zitzler1999multiobjective} and stands as one of the fundamental unary indicators for assessing the quality of a Pareto front approximation set. Notably, this indicator possesses a distinct advantage because it does not necessitate prior knowledge of the Pareto front. Maximizing the HV can yield a Pareto front approximation set that is both highly qualified and diverse \cite{yang2019efficient}. The function of the HV indicator is to measure the size of the subspace dominated by $\mathcal{P}$, which is bounded below by a reference point ${\bf r}$ and defined as 

$$HV(\mathcal{P})=\lambda_{m}\left(\cup_{{\bf y}\in \mathcal{P}}\left[{\bf r},{\bf y}\right]\right)$$

where $\lambda_{m}$ represents the Lebesgue measure of an $m$-dimensional subspace bounded by $\left[{\bf r},{\bf y}\right]$. Following the process described in \cite{yang2019efficient}, we select the reference point such that it is dominated by all elements in the Pareto front approximation set.
%where $\lambda_{m}$ is the Lebesgue measure of a $m$-dimensional subspace bounded by $\left[{\bf r},{\bf y}\right]$. As suggested in \cite{yang2019efficient}, the reference point can be selected so that it is dominated by all the elements in a Pareto front approximation set.

Given an approximation of the Pareto front $P$, the improvement in hypervolume resulting from adding a new solution vector ${\bf y}$ is defined as $HVI\left({\bf y}, P\right)=HV\left(P\cup {\bf y}\right)-HV\left(P\right)$. Thus, the Expected Hypervolume Improvement (EHVI) generalizes the concept of Expected Improvement (EI) to multi-objective optimization by identifying the solution ${\bf y}$ that maximally increases the volume of the subspace dominated by $P$ \cite{yang2019efficient}. Furthermore, like EI in single-objective settings, EHVI supports parallel candidate generation ($q$EHVI) and gradient-based acquisition optimization \cite{daulton2021parallel}.

%Accordingly, given an approximation of the Pareto front $P$, the improvement in HV due to the incorporation of a new vector of solutions ${\bf y}$ is given by $HVI\left({\bf y}, P\right)=HV\left(P\cup {\bf y}\right)-HV\left(P\right)$. Therefore, the Expected hypervolume improvement (EHVI) extends the concept of EI to Multi-objective optimization settings by looking for the solution ${\bf y}$ that expands the volume of the subspace dominated by $P$ the most \cite{yang2019efficient}. Moreover, similar to the single solution EI, EHVI supports parallel candidate generation ($q$EHVI) and gradient-based acquisition optimization \cite{daulton2021parallel}. 

However, the standard EHVI has certain limitations, such as assuming noise-free observations and the exponential complexity of its batch variant, $q$EHVI, which restricts large-batch optimization \cite{daulton2021parallel}. In cases of noisy observations, a variant known as Noisy Expected Hypervolume Improvement (NEHVI) accounts for uncertainty in the function values at observed points and can also be adapted for parallel settings as $q$NEHVI \cite{daulton2021parallel}. Estimating $q$NEHVI involves solving multiple multivariate integrals, and despite several exact and approximate methods in the literature, their complexity for large $m$ often leads to the use of numerical methods based on Monte Carlo (MC) simulations \cite{yang2019multi}. Consequently, both $q$EHVI and $q$NEHVI rely on setting an appropriate number of MC samples to accurately estimate the acquisition function during optimization.

%Still, the standard EHVI has some limitations, including the assumption that observations are noise-free and the exponential scaling of its batch variant, $q$EHVI, which precludes large-batch optimization \cite{daulton2021parallel}. When there are noisy observations, there is a variant of EHVI called Noisy expected Hypervolume improvement (NEHVI), which integrates over the uncertainty in the function values at the observed points and can also be extended to parallel settings ($q$NEHVI) \cite{daulton2021parallel}. The estimation of $q$NEHVI requires solving multiple multi-variate integrals, and even though there are several exact and approximate proposals in the literature, due to their complexity for large $m$, the most widely extended approach is to use numerical methods based on Monte Carlo (MC) \cite{yang2019multi}. Therefore, both $q$EHVI and $q$NEHVI are required to set up the number of MC samples to estimate the AF correctly during its optimization process. 

\subsection{Multi-objective Regionalized Bayesian Optimization}

As discussed before, MORBO optimizes diverse parts of the global Pareto frontier in parallel using a coordinated set of local TRs and defines a set of heuristic rules to expand, decrease, or even reset the TRs according to their success/failure at providing new candidate solutions. The implementation of the MORBO strategy requires setting seven new hyperparameters: 

\begin{itemize}
    \item The number of Trust Regions $n_{TR}$.
    \item The number of Thomson samples $n_{TS}$ for sampling the posterior distribution of every TR's GP.
    \item The initial TR length $L_{init}$.
    \item The maximum and minimum TR lengths, denoted as $L_{max}$ and $L_{min}$ respectively.
    \item The success/failure thresholds, $\tau_{succ}$, $\tau_{fail}$ respectively.  
\end{itemize}

To initialize the TR, MORBO chooses the centers of the TR as the points in $P$ with maximum hypervolume contribution (HVC), i.e., given a reference point $\bf{r}$, HVC of a solution $\bf{y}$ on $P$ is the reduction in HV if that point were to be removed. MORBO selects TR centers based on their HVCs in a sequential greedy fashion, excluding points already selected as the center for another TR \cite{daulton2022multi}. The original MORBO approach uses the sample with the lowest amount of total constraint violation as a center in case there is no feasible point. However, considering that the addressed swine diet formulation problem has 28 zootechnical and nutritional restrictions, the feasible space is likely a considerably shrunken version of the search space, and probably no feasible solution can come up at random. Therefore, in this work, the initial set of solutions is obtained by optimizing a constant function that must satisfy the whole set of constraints. The length of the $j$-th TR is initialized as $L_j=L_{init}$, and local GPs are trained per TR using observations within a hypercube with an edge length of $2L_j$.

Unlike the MOBO process, MORBO does not need to optimize any AF. However, MORBO requires executing two processes: finding the new query point ${\bf x}_{n+1}$ and updating TRs. Table \ref{tab:MORBO_processes} summarizes the two components of MORBO. The process of finding new query points from previously set TRs (left column in Table \ref{tab:MORBO_processes}) is straightforward; nevertheless, to understand the TRs updating process fully, it is necessary to consider that MORBO maintains success and failure counters that record the number of consecutive samples generated from the TR that improved or failed to improve (respectively) the objective \cite{daulton2022multi}. When the success counter exceeds a predetermined $\tau_{succ}$, the TR length is increased to $min\{2L,L_{max}\}$ and the counter is reset to zero. Oppositely, after $\tau_{fail}$ consecutive failures the TR length is set to
$L/2$, and the failure counter is set to zero. If the length $L_j$ drops below a minimum edge length $L_{min}$, the TR is terminated, and a new TR is initialized. The center of the reset TR is obtained following a hypervolume scalarization procedure based on \cite{zhang2020random} that ensures TRs are initialized in diverse parts of the objective space and, according to \cite{daulton2022multi}, yields a global optimization performance guarantee.

\begin{table}[]
    \centering
    \caption{Processes involved during functions optimization using MORBO}
    \label{tab:MORBO_processes}
    \begin{tabular}{p{8cm}|p{8cm}}\hline
     \textbf{Finding new query points} &  \textbf{Updating TRs}\\\hline
     Given an initial set $\mathcal{D}_{{\bf f}}$, and initialized TRs,
     \begin{enumerate}
         \item For each TR:
         \begin{itemize}
             \item Get $n_{TS}$ samples from the corresponding GP posterior distribution.
             \item Filter out those candidates that do not satisfy the restrictions.
             \item Estimate the HVI of feasible solutions
         \end{itemize}
         \item Find the $q$ candidate solutions among all the TRs with the highest HVI. Once sampled from the vector-valued target function $\{{\bf f}({\bf x}_{n+1}), {\bf f}({\bf x}_{n+2}), \cdots, {\bf f}({\bf x}_{n+q})\}$, they can be added to the history of observations $\mathcal{D}_{{\bf f}}=\mathcal{D}_{ {\bf f}} \cup \{({\bf x}_{n+i}, {\bf y}_{n+i})\}_{i=1}^q$.
     \end{enumerate}
        & Once the previous process ends, 
        \begin{enumerate}
            \item Update the TRs by adding the new samples when they fall into the TR area.
            \item For each TR:
         \begin{itemize}
         \item Increment success/failure counters and lengths $L_j$.
         \begin{itemize}
         \item If the $L_j < L_{min}$, the TR is reset, getting the center from $\mathcal{D}_{{\bf f}}$, defining $L_j = L_{init}$, assigning samples and training a new local GP.
         \end{itemize}
         \item Update center to the locally available point with maximum HVC.
         \item Train a local GP with observations within a hypercube with an edge length of $2L_j$.
         \end{itemize}
         
        \end{enumerate}  \\\hline
    \end{tabular}
\end{table}

\subsection{Optimality evaluation}\label{sec:optimality}

The quality of a solution set in multi-objective optimization problems is typically assessed by evaluating the depiction of the Pareto front through four primary aspects. These aspects are:

\begin{itemize}
    \item Convergence: This measures how closely the solution set approaches the Pareto front, which, in this case, is measured through HV.
    \item Cardinality: This assesses the number of elements within the solution set.
    \item Diversity: This is crucial for understanding how well the set of non-dominated solutions spans the Pareto front, providing decision-makers with a range of different and valuable alternatives.
\end{itemize}

Given a set of observations $\mathcal{D}_{\bf f}$ at an iteration $k$, Cardinality $C_{\mathcal{D}_{\bf f}} = t$ is the number of non-dominated solutions in $\mathcal{D}_{\bf f}$. A large $C_{\mathcal{D}_{\bf f}}$ means that the method is able to find many non-dominated solutions to approximate the Pareto front set $P$.

Although the HV indicator is associated with both the size and spread of the solution set (since sets with more non-dominated solutions naturally cover a larger space), it does not guarantee a uniform distribution of the solutions. As a result, diversity is often measured indirectly by combining two indicators: spread, which reflects the extent of the solution set's coverage, and uniformity, which evaluates how evenly the solutions are distributed \cite{li2019quality}.
%While the HV indicator is linked to the cardinality and spread of the solution set, since sets with more non-dominated elements inherently cover a larger space, the HV indicator alone does not ensure the uniform distribution of the solution set. Therefore, usually, the diversity is measured indirectly through the combination of two indicators: the spread, reflecting the coverage of the solution set, and uniformity, assessing the even distribution of the set \cite{li2019quality}. 
However, as indicated in \cite{cai2018diversity}, when the many solutions in $P$ are located on the boundaries of $\mathcal{P}$, either the spread or uniformity indicator fails to give the correct evaluation of diversity. Therefore, alongside the HV metric, the evaluation of the Pareto front approximation provided by MOBO and MORBO approaches will include the diversity indicator based on reference vectors (DIR) as proposed in \cite{cai2018diversity}. DIR utilizes a set of $u$ reference vectors $V=\{\nu_{1}, \nu_{2}, \dots \nu_{u}\}$ that are uniformly generated to cover the solution space $\mathcal{Y}$. A coverage vector ${\boldsymbol{\upsilon}}=[\upsilon_{1},\upsilon_{2},\dots, \upsilon_{t}]^{T}$ is then created to count the number of times each solution in $P$ is the closest to each reference vector in $V$. DIR is defined as:
\[
DIR(P)=\dfrac{\sqrt{\frac{1}{t}\sum_{i=1}^{t}(\upsilon_{i}-\text{mean}({\boldsymbol{\upsilon}}))^{2}}}{\frac{u}{t}\sqrt{t-1}}
\]
Lower DIR values mean better diversity of the solutions in $P$.

\section{Experiments and results}\label{sec:exper_res}

\subsection{Experimental setup}\label{sec:exper_setup}

The experiments are organized in four phases: I) Comparative analysis of convergence between MOBO and MORBO processes; II) Selection of MORBO hyperparameters; III) Assessment of the quality of MORBO solutions compared to the results achieved by \cite{pena2009multiobjective, guribe2024}. IV) evaluation of the effects of parallel configurations of MORBO to speed up the optimization process. In all experiments, the surrogate GP models used by both MOBO and MORBO methods employ an isotropic Matérn kernel, with its hyperparameters estimated by maximizing the log-likelihood based on the data. The Matérn kernel is a generalization of the exponentiated quadratic kernel, introducing an additional parameter that controls the smoothness of the resulting function. This kernel is particularly favored for high-dimensional problems \cite{williams2006gaussian}.
%For all the experiments, the surrogate GP models trained by MOBO and MORBO methods use an isotropic Matérn Kernel, whose hyperparameters are estimated by maximizing the log-likelihood with respect to the data. Matérn kernel is a generalization of the exponentiated quadratic that adds an additional parameter controlling the smoothness of the resulting function, and it is preferred for high dimensional problems \cite{williams2006gaussian}.

The MOBO parameters (number of initial samples, MC samples, and iterations) are set according to the findings in \cite{guribe2024}. During phase I, $n_{TS}$ is evaluated on the grid $\{512, 1024, 2048, 4096\}$ to test its impact on the convergence properties of the method in swine diet design and to identify the minimum number needed to achieve convergence before reaching the Pareto set. To analyze the convergence properties of MORBO, the maximum number of iterations is set to $k=150$, which is three times the value identified in \cite{guribe2024} for MOBO to achieve an HV plateau. However, for the sake of comparison, from phase II onwards, $k$ is set to 50 to ensure fair competition with MOBO, and the quality of the solutions is compared between MOBO and MORBO under similar conditions. The number of initial samples in MORBO is set to the same value used by MOBO (50) \cite{guribe2024}. The values of $n_{TR}$ are explored within the set $\{2, 5 ,8\}$, as larger values increase the variance of HV values obtained by MORBO. Additionally, the initial length of the trust regions $L_{init}$ is examined among the values $\{0.1, 0.2, 0.4\}$, which cover a large percentage of the possible variables range.

Considering that the solution vectors are subject to a sum-to-one constraint (input variables are normalized), the success/failure thresholds and maximum and minimum values of the lengths of the TRs were set automatically according to the methods defined in \cite{daulton2022multi}. Given the stochastic nature of some parts of MOBO and MORBO, all the experiments are repeated 30 times using different seeds to evaluate the stability of the results.  Subsequently, during phases III and IV, the best set of hyperparameters is used to get the Pareto set and Pareto front approximation of the multiobjective swine diet design problem described in Section \ref{sec:problem}. The quality of the Pareto front approximation is evaluated in terms of HV, $C_{\mathcal{D}_{\bf f}}$, and DIR. The number of reference vectors $u$ required to estimate DIR was chosen as 78 after simple experimentation to find out how many reference vectors are required to get a stable metric value regarding the solution space of the diet design problem described in section \ref{sec:problem}. Since one of the main concerns about MOBO is its limitation in exploring the feasible space, different analyses are included to illustrate comparatively how MOBO and MORBO perform in this respect. The distance between consecutive solutions in the ingredients, nutritional, and target spaces is analyzed along with visual representations of the distribution of the Pareto set and Pareto front approximation. In order to graphically represent the distribution of solutions in the Pareto set and Pareto front, a Uniform Manifold Approximation and Projection (UMAP) \cite{mcinnes2018umap} technique is employed to project the 17-dimensional and 3-dimensional vectors of solutions, respectively, into a two-dimensional space. UMAP is a manifold learning technique widely used for visualizing the structure of high-dimensional data in low-dimensional spaces that, unlike other techniques, aims to capture both the global and local structure of the data and optimizes an embedding that preserves the distances and connectivity of the samples in their original space. The projection is complemented with information regarding the uncertainty of the GP model when every point was selected as a candidate solution. This representation allows us to understand better the behavior of MOBO and MORBO methods regarding exploration/exploitation characteristics in the context of the swine diet design problem.

Later, similar to the analysis presented in \cite{guribe2024}, the quality of the solutions provided by MORBO is analyzed in comparison to those provided by MOBO, and also the single solution provided by the MFP method proposed in \cite{pena2009multiobjective}. This evaluation is performed both in terms of dominance and also according to the level of improvement per objective measured using the distance metric given by
\begin{equation}\label{eq:dist}
    d_{{y}_j}=\dfrac{y_{j}-\text{MFP}_{j}}{\max_j-\min_j} \text{ with } j\in \left\lbrace1, 2, \dots m\right\rbrace 
\end{equation}

where $y_j$ and $\text{MFP}_j$ represent the values obtained for the $j$-th objective by the MOBO/MORBO and MFP solutions, respectively. Additionally, $\max_j$ and $\min_j$ denote the maximum and minimum values observed during the simulations for the $j$-th objective. In other words, the difference between the two objective values is normalized based on the observed range of that objective across feasible solutions.
%where $y_j$ and $\text{MFP}_j$ are the values obtained for $j$-th objective by MOBO/MORBO and MFP solutions, respectively. Besides, $\max_j$ and $\min_j$ are the maximum and minimum values observed during simulations for the objective $j$. In other words, the difference between the two objective values is normalized with respect to the observed range of that objective in feasible solutions.

Lastly, during phase IV, the quality of the solutions is also analyzed for parallel settings when $q=2,3$, paying special attention to HV levels and the capacity of MORBO and MOBO to provide dominant solutions compared to that of the MFP method.  

\subsection{Results}

\begin{figure}[H]
    \centering
    \includegraphics[scale=0.4]{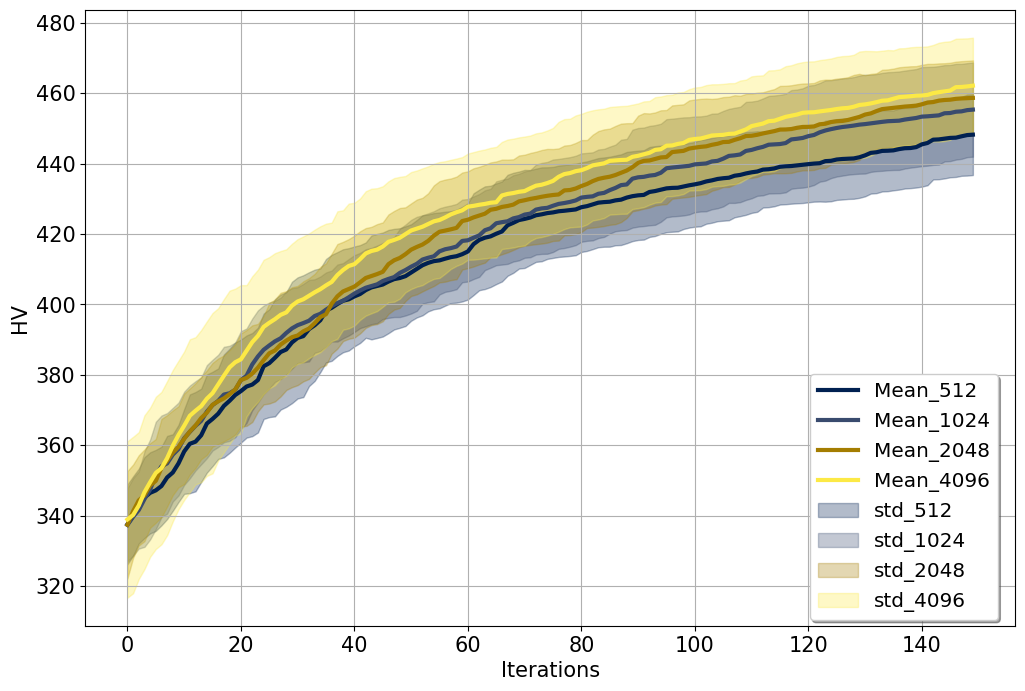}
    \caption{Convergence of MORBO process for HV values across iterations and for different variations of $n_{TS}$. Solid lines represent the mean value for each $n_{TS}$, and shaded areas represent the mean $\pm$ the standard deviation per $n_{TS}$.}
    \label{fig: fig1}
\end{figure}

Figure \ref{fig: fig1} shows the convergence properties of the MORBO process measured through the improvement of HV across iterations and for different $n_{TS}$ values. Solid lines represent the average hypervolume achieved by each configuration, and the shaded areas indicate the standard deviation over the iterations. The graph shows an increasing trend across all configurations. Additionally, it is evident that although convergence has not yet been achieved, the rate of growth acceleration is decreasing. Furthermore, it can be observed that as the number of Thompson samples ($n_{TS}$) increases, the convergence of MORBO improves slightly.

From figure \ref{fig: fig1} is possible to observe that for values of $n_{TS}$ greater than $1024$, the average HV value obtained by MORBO approaches $460$ when $150$ iterations are reached, and the optimization process still do not converge. According to the results presented in \cite{guribe2024}, the MOBO method achieves average HV values around $466.92$ with only 50 iterations, which is a much better result in terms of convergence rate. This behavior could be explained by the fact that MOBO and the NEHV AF promote the search process to explore places closer to the boundaries of the feasible space during the early stages of the optimization process, which makes the HV increase at a faster rate during the first iterations. However, as we will see in the following results, such a fast exploration of the search space limits does not guarantee either a better coverage of the search space and the Pareto front approximation or more diverse solutions in the Pareto set.

\begin{figure}
    \centering
    \includegraphics[scale=0.35]{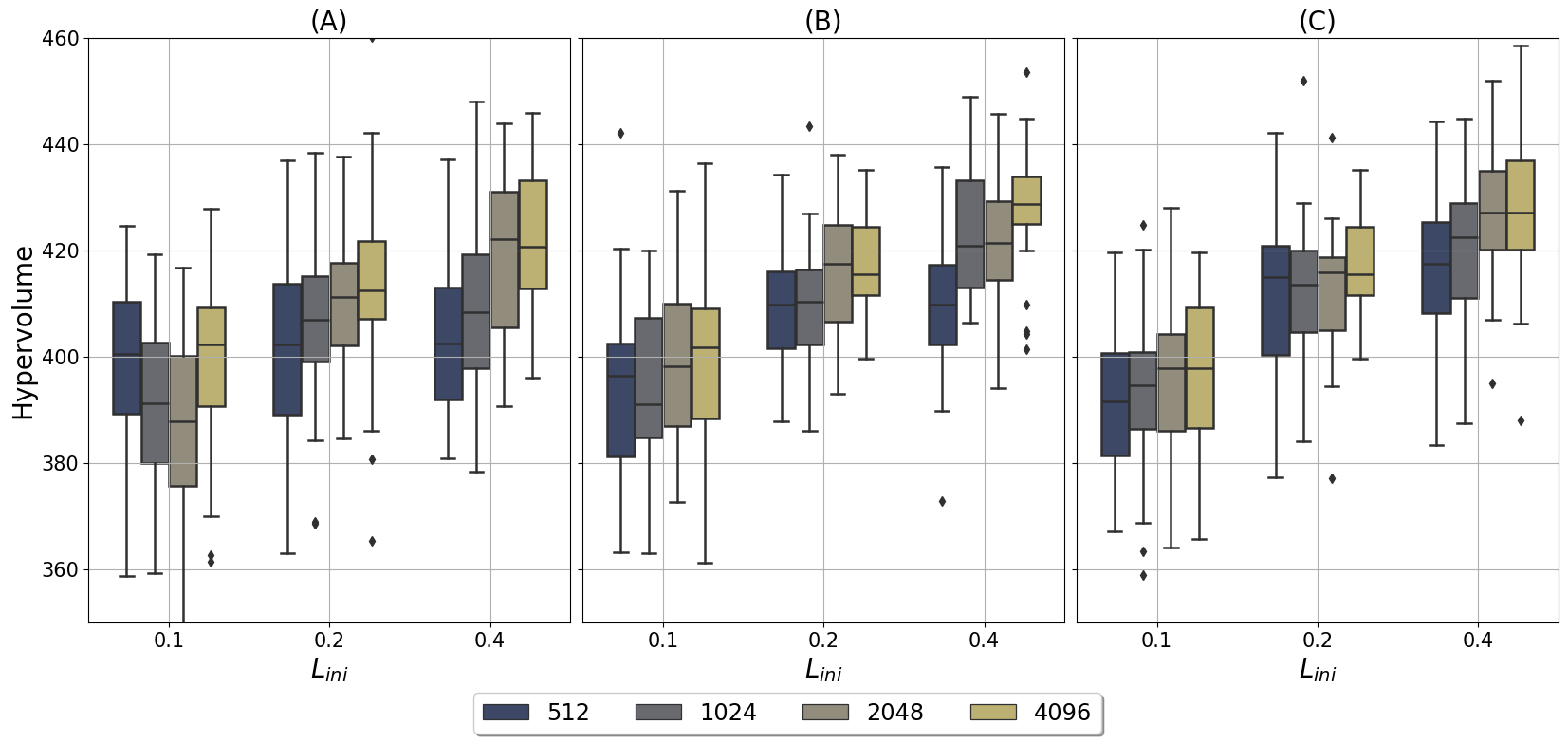}
    \caption{Boxplots of HV obtained by MORBO for different hyperparameter configurations during 30 executions. (A) shows boxplots for $n_{TR} = 3$; (B) shows boxplots for $n_{TR} = 5$; and (C) shows boxplots for $n_{TR} = 8$.}
    \label{fig:hyper}
\end{figure}

Figure \ref{fig:hyper} shows box plots of HV values obtained by MORBO during multiple executions of different hyperparameter configurations. According to the results, the most substantial effects are associated with $L_{init}$ and $n_{TS}$ and less with $n_{TR}$. In subfigures (A), (B), and (C) of figure \ref{fig:hyper}, the most noteworthy growth trend depends on the values of $L_{init}$. The largest HV values were obtained when $L_{init}=0.4$. Values larger than 0.4 were also tested, but no improvements were observed, i.e., for MORBO configuration with $n_{init}=0.8$, and different  $n_{TR}$ values, the average HV did not surpass $423.53$. The largest average HV in figure \ref{fig:hyper} corresponds to a MORBO configuration with $n_{TS}=4096$ and $n_{TR}=8$, which achieves an HV of 428.68. However, this setup also exhibits high instability with a standard deviation of 15.32. The HV for $n_{TS}=4096$ and $n_{TR}=5$ achieves a value of 427.92 with a considerably smaller standard deviation of 11.62, which will be considered the best for the subsequent analyses.

\begin{figure}
    \centering
    \includegraphics[scale=0.25]{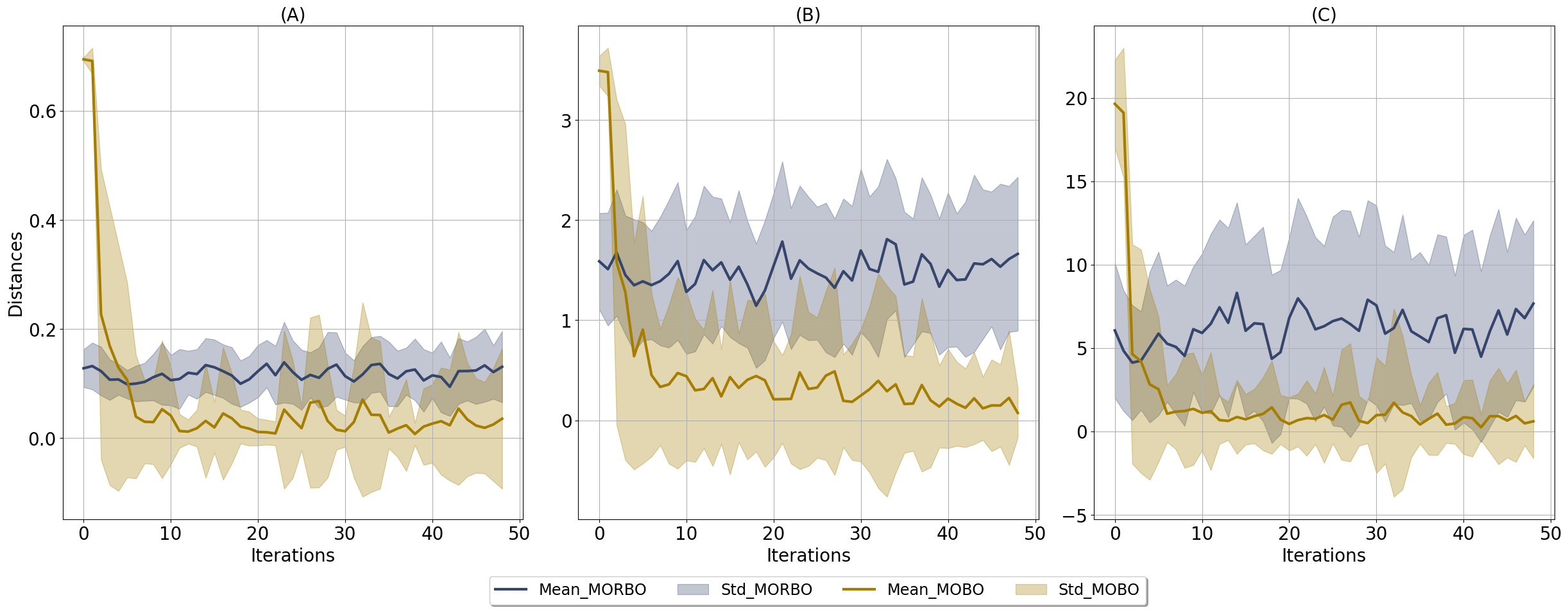}
    \caption{Figure (A) illustrates the average distances between consecutive solutions in the ingredient vector space. Figure (B) presents the average distances between consecutive solutions estimated in the nutrient vector space, while Figure (C) displays the average distances between consecutive solutions in the objective space. The distances are computed over 50 iterations for all figures and for both the MOBO and MORBO methods.}\label{fig:explore1}
\end{figure}

Regarding exploration, Figure \ref{fig:explore1} displays the average Euclidean distance between consecutive solutions provided by MOBO and MORBO, along with the standard deviation throughout the optimization process. This analysis was conducted in three different spaces: input (ingredients), constraints (nutrients), and outputs (objectives). From Figure \ref{fig:explore1}, it is evident that at the start of the optimization process, MOBO promotes the exploration of regions distant from the initial set of solutions, where model uncertainty is high. However, as the optimization progresses, its exploration significantly decreases. In contrast, MORBO does not explore regions in the input space too far from known solutions, but it maintains a more consistent level of exploration across iterations, which is reflected in the three spaces involved.

\begin{figure}[ht!]
\setlength\tabcolsep{2pt}%%
\centering
\begin{tabular}{cccc}
 & \textbf{Pareto set} & \textbf{Pareto front} & \\
 \raisebox{5.5\normalbaselineskip}[0pt][0pt]{\rotatebox[origin=c]{90}{MOBO}} &
 \includegraphics[width=0.33\textwidth]{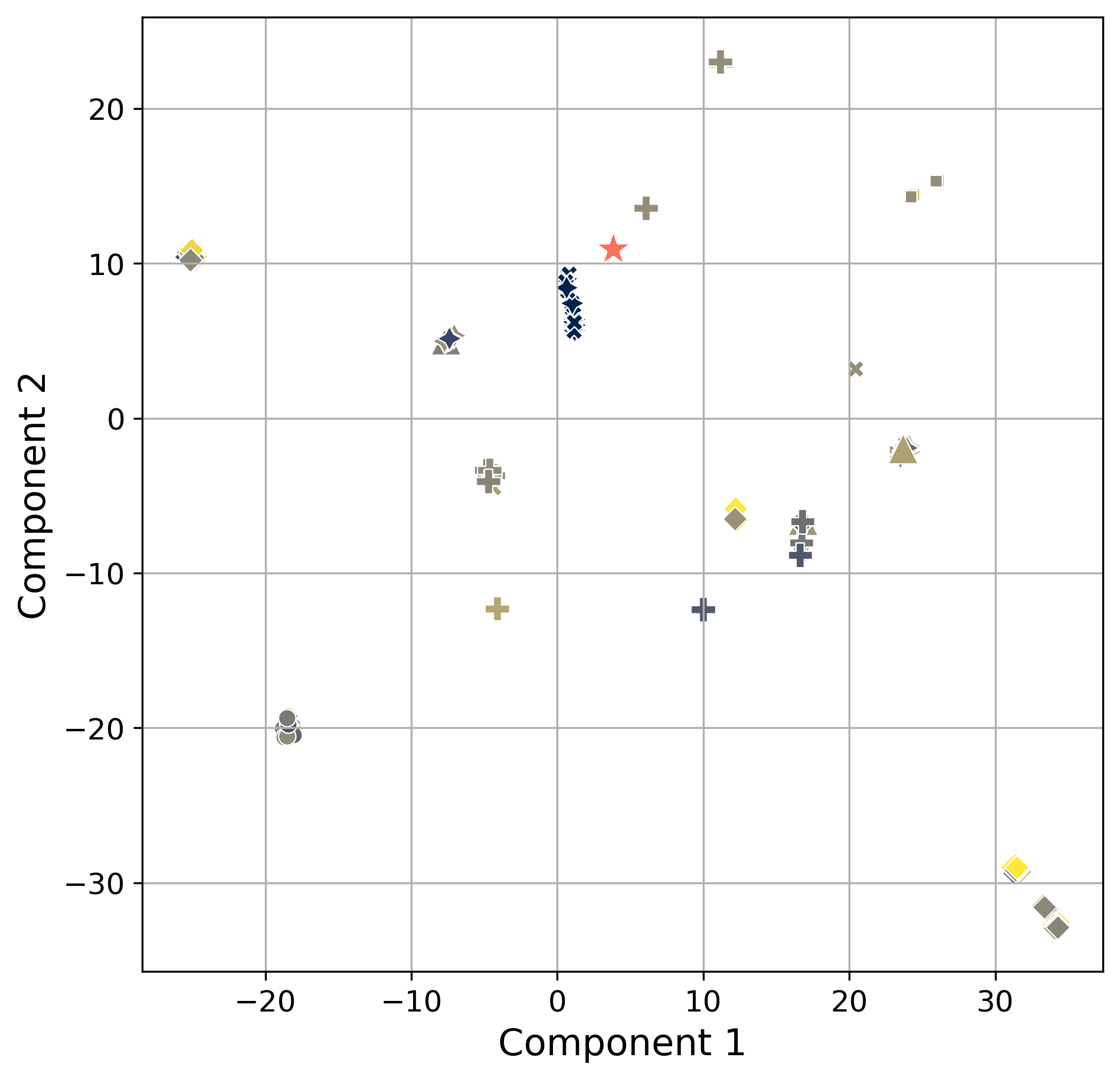} &
 \includegraphics[width=0.33\textwidth]{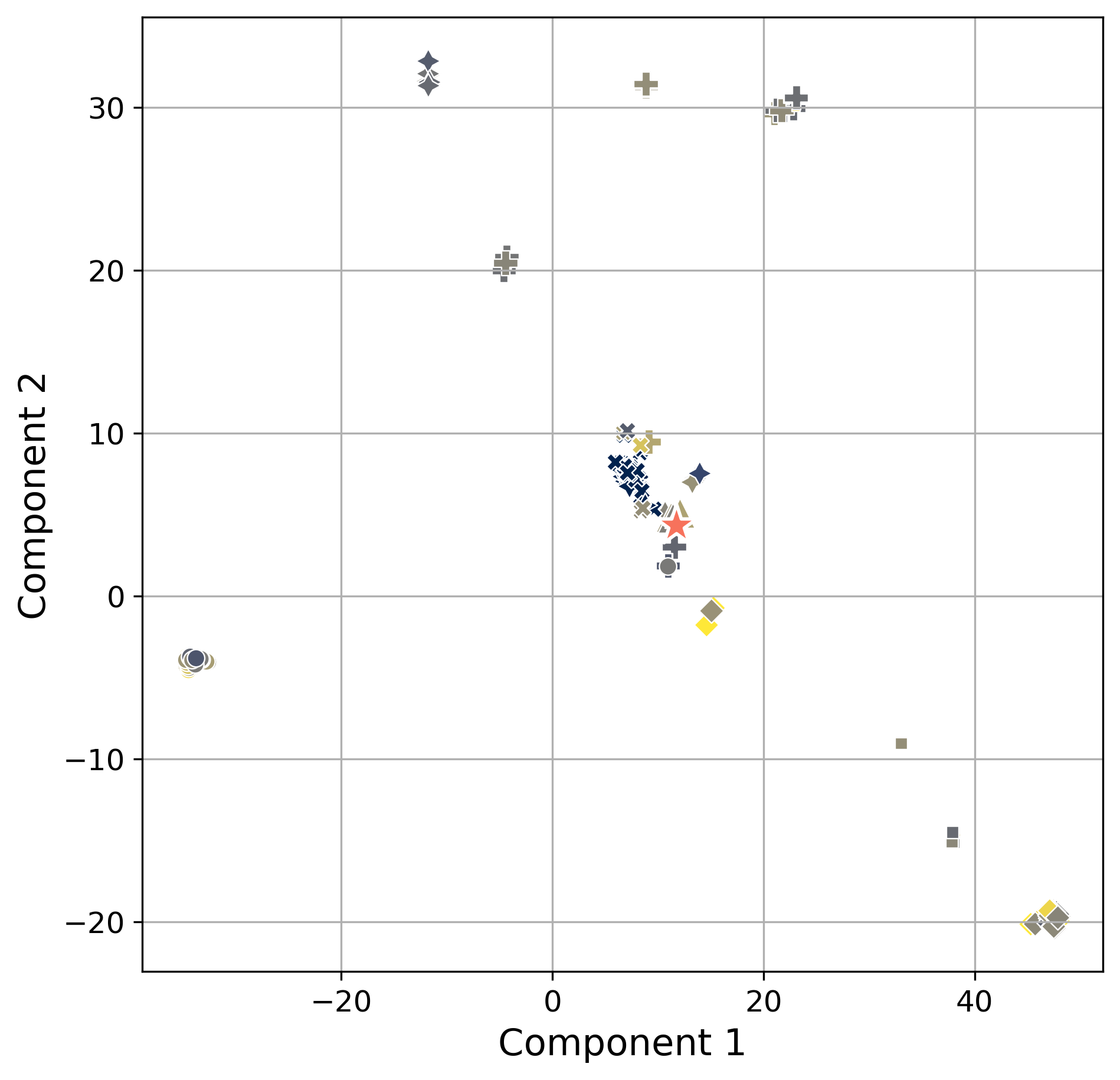} &
 \multirow{2}{*}[8.5\normalbaselineskip]{\includegraphics[width=0.07\textwidth]{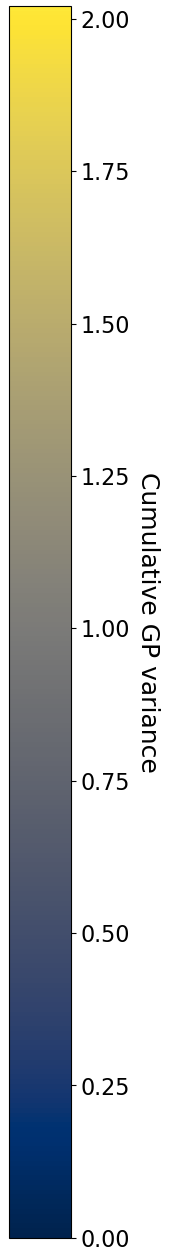}} \\
 \raisebox{7\normalbaselineskip}[0pt][0pt]{\rotatebox[origin=c]{90}{MORBO}} &
 \includegraphics[width=0.33\textwidth]{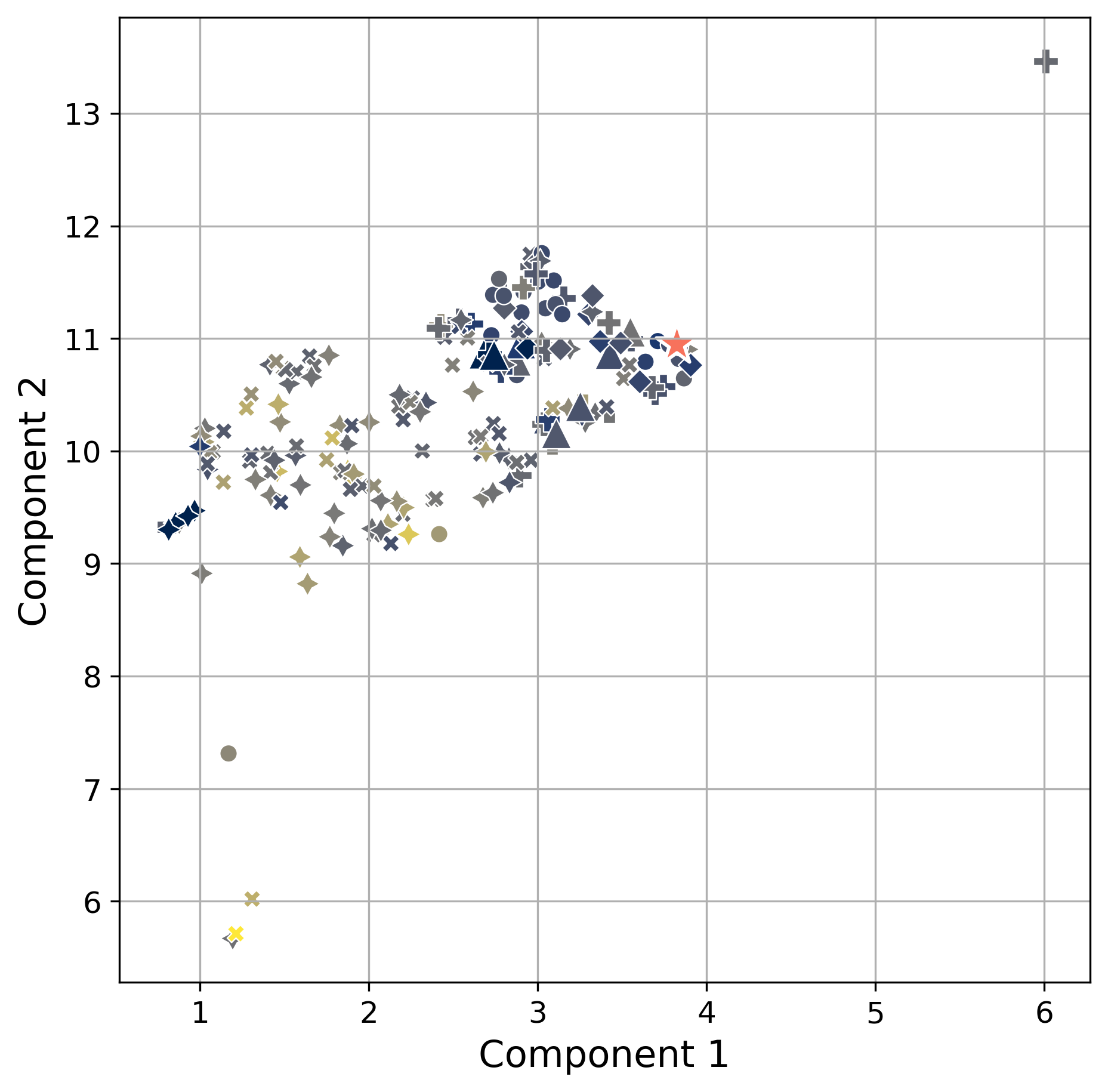} &
 \includegraphics[width=0.33\textwidth]{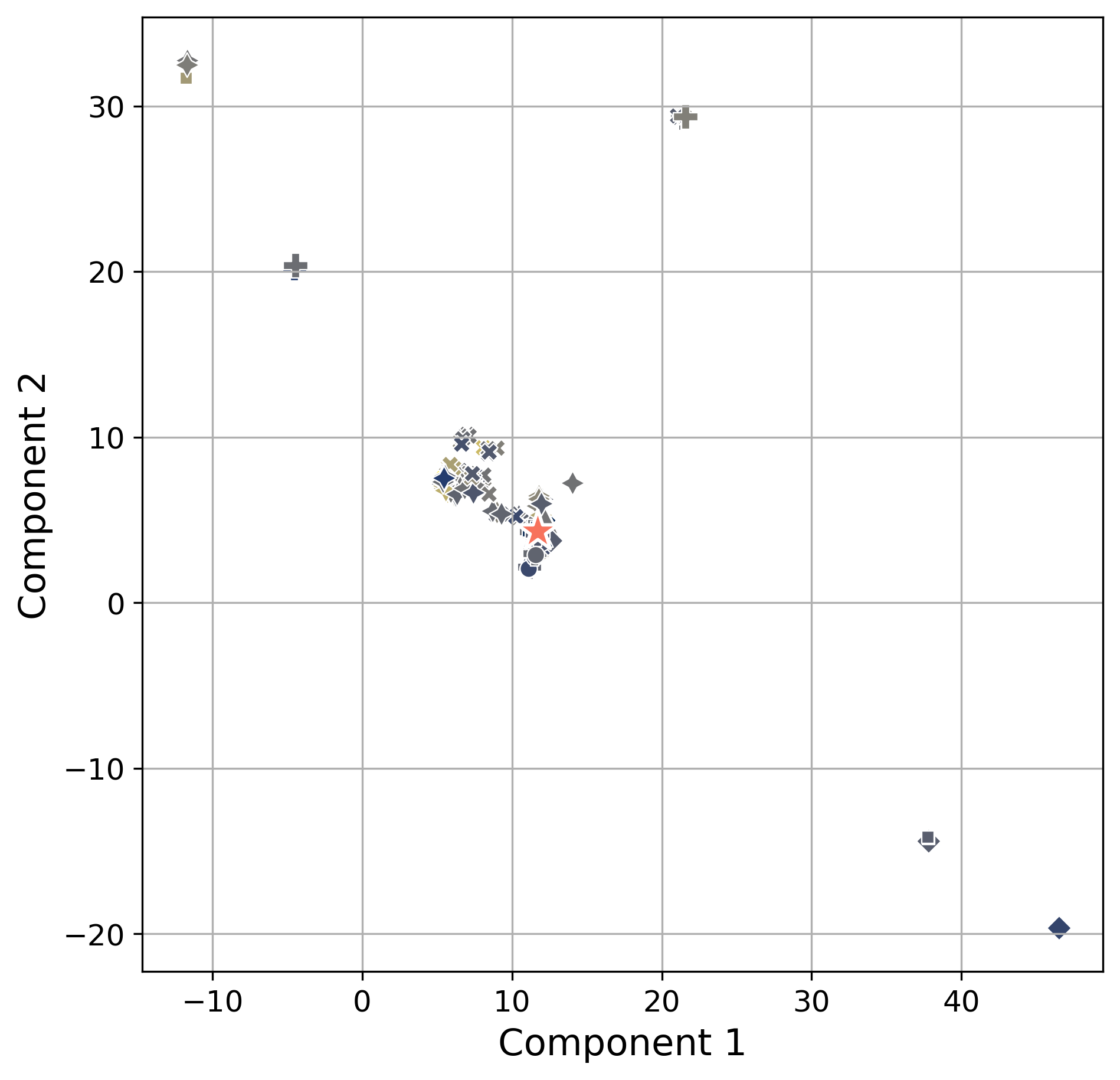} & \\
 \multicolumn{4}{c}{\includegraphics[width=0.5\textwidth]{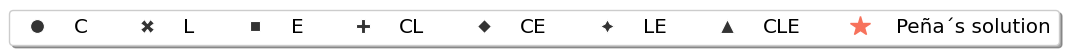}}
\end{tabular}
\caption{UMAP projections of Pareto sets and Pareto fronts obtained by MOBO and MORBO. The solution provided by \cite{pena2009multiobjective} is also included for the sake of comparison. The color bar represents the cumulative variance for the three objectives obtained from their corresponding GP surrogate models at the time when each of the solutions was suggested as the new query point during the BO process. Each marker corresponds to solutions that outperform the one obtained in \cite{pena2009multiobjective} for each objective: cost (C), Lysine (L), energy (E), pairs of objectives: Cost-Lysine (CL), Cost-Energy (CE), Lysine-Energy (LE), and all three objectives combined, Cost-Lysine-Energy (CLE).}\label{fig:UMAP}
\end{figure}

\begin{figure}
    \centering
    \includegraphics[scale=0.45]{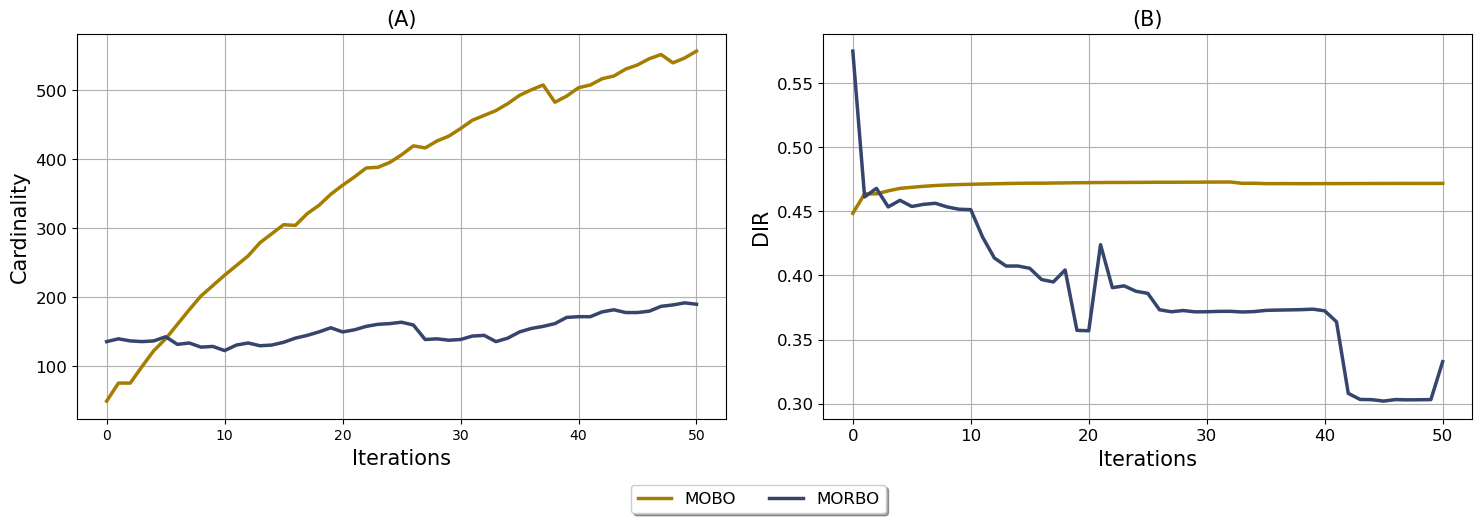}
    \caption{Figure (A) shows the cumulative cardinality in MOBO and MORBO over 50 iterations during the 30 executions. Figure (B) shows DIR quality indicators for the cumulative set of Pareto front approximation over 50 iterations and during the 30 executions.}
    \label{fig:DIR}
\end{figure}

Alternatively, Figure \ref{fig:UMAP} presents the two-dimensional UMAP projection of all non-dominated solutions obtained from the MOBO and MORBO methods during the 30 runs of the experiment. The figure includes projections of both the Pareto set and the Pareto front. By comparing the two left-sided plots of Figure \ref{fig:UMAP} (Pareto sets), it becomes evident that MORBO's exploration results in a more uniform coverage of the feasible space. Given that UMAP preserves the global structure of the data, the dynamic range of the MOBO solutions confirms that MOBO primarily explores zones closer to the boundaries of the search/feasible space, as \cite{wang2020learning} suggested, while MORBO seeks solutions within the inner areas of the search/feasible space. Remarkably, as can be seen in figure \ref{fig:DIR}(A), the value of $C_{\mathcal{D}_{\bf f}}$ for MOBO is 2.5x larger than that of MORBO, but MOBO's DIR value is considerable higher (see figure \ref{fig:DIR}(B)), which means that most of the solutions in the $P$ provided by MOBO are so similar than they appear overlapped in the UMAP projection of Figure \ref{fig:UMAP}. Therefore, these results suggest that, although during the first iterations MOBO explores regions with very high uncertainty and close to the boundaries of the search space, most of the new solutions come from known regions with almost no uncertainty and whose difference from previous solutions is marginal. This is also supported by the results in figure \ref{fig:DIR}; while the $C_{\mathcal{D}_{\bf f}}$ increases along iterations, the DIR indicator keeps almost contact, so the new nondominated solutions added to $P$ does not improve coverage of the Pareto front. The fact that MOBO finds a larger number of solutions close to the boundaries of the search spaces is the reason why its HV values are larger than those of MORBO. On the contrary, MORBO promotes a better balance between exploration and exploitation during the optimization process, which is reflected in the UMAP projections of the Pareto set in Figure \ref{fig:UMAP}, and also in the decreasing values of DIR as the number of iterations reaches 50. Interestingly, by analyzing the right-sided plots of Figure \ref{fig:UMAP} (Pareto fronts), the mapping between the input and output spaces seems to be surjective since less diversity is observed, although the results shown in Figure \ref{fig:DIR} confirm that also the diversity of $P$ is better for MORBO than MOBO. 

Similar to the analysis performed in \cite{guribe2024}, Table \ref{tab:solution_comp}
shows the percentage of solutions obtained by the MORBO configuration, which improve one, two, or three MFP objectives \cite{pena2009multiobjective}.
Table \ref{tab:solution_comp} uses a similar notation as Figure \ref{fig:UMAP} for solutions that outperform that obtained in \cite{pena2009multiobjective} for each objective: cost (C), Lysine (L), energy (E), pairs of objectives: Cost-Lysine (CL), Cost-Energy (CE), Lysine-Energy (LE), and finally, all three objectives combined, Cost-Lysine-Energy (CLE). Table \ref{tab:solution_comp} also shows the number of executions that yield at least one solution dominating that provided by the MFP method for each of the optimization methods and for different numbers of iterations. It can be seen that by going through the iterations, MORBO is progressively improving while MOBO, after 30 iterations, is unable to find additional solutions dominating MFP. These results suggest that MORBO's exploration of the search space is more efficient even though its HV values do not reach those of MOBO. Moreover, the surjective characteristic of the mapping between the input and output spaces in the swine diet design problem could explain that even though the cardinality of $P$ is larger for MOBO, the poor diversity of the solutions prevents MOBO from finding more solutions dominating that of the MFP. 

\begin{table}[H]
\centering
\caption{Percentages of solutions that improve 0, 1, 2, or 3 objectives compared to MFP \cite{pena2009multiobjective}. The number of executions (runs) where the optimization process was able to find solutions dominating that of the MFP is also included.}
\label{tab:solution_comp}
\resizebox{\textwidth}{!}{\begin{tabular}{clrrrrrrrrr}
\hline
& $k$ &  $\emptyset$($\%$) & C($\%$)& L ($\%$)& E($\%$)& CL ($\%$)& CE ($\%$)& LE ($\%$)& CLE ($\%$) & $\#$ run ($\%$)\\
\hline
\multirow{5}{*}{\rotatebox[origin=c]{90}{\bf MOBO}} & 10 &  0.69 &    29.15 &    74.45 &    31.21 &      5.14 &     18.70 &     11.85 &       0.19 &     1 (3.33) \\
& 20 &  0.59 &    36.58 &    68.43 &    37.17 &      7.26 &     19.88 &     16.04 &       0.40 &     3 (10.00) \\
& 30 &  0.52 &    41.39 &    67.14 &    38.89 &     11.21 &     19.91 &     17.27 &       0.43 &     4 (13.33)\\
& 40 &  0.46 &    44.03 &    67.37 &    38.54 &     14.24 &     19.33 &     17.20 &       0.38 &     4 (13.33) \\
& 50 &  0.43 &    45.39 &    67.68 &    38.02 &     16.03 &     18.86 &     16.98 &       0.36 &     4 (13.33) \\
\hline
\multirow{5}{*}{\rotatebox[origin=c]{90}{\bf MORBO}} & 10 &  0.10 &    12.33 &    95.31 &    21.28 &      8.00 &      1.76 &     20.43 &       1.19 &     8 (26.67) \\
& 20 &  0.10 &    17.67 &    90.69 &    22.65 &      9.18 &      2.62 &     20.90 &       1.59 &    11 (36.67) \\
& 30 &  0.09 &    21.82 &    87.22 &    25.45 &     10.00 &      3.59 &     22.66 &       1.70 &    13 (43.33) \\
& 40 &  0.08 &    24.90 &    84.79 &    27.49 &     10.66 &      4.00 &     24.45 &       1.83 &    15 (50.00) \\
& 50 &  0.07 &    26.12 &    83.78 &    28.65 &     10.80 &      4.55 &     25.09 &       1.82 &    17 (56.67) \\ \hline

\end{tabular}}
\end{table}

It is worth mentioning that, in four out of the seventeen runs where MORBO found solutions dominating that of the MFP, the process yielded multiple solutions, with one run achieving up to three solutions that enhanced all three objectives. Moreover, from Table \ref{tab:solution_comp}, it can be seen that after 40 iterations, with the MORBO method, more than $99\%$ of the solutions in the Pareto front approximation are not dominated compared to the MFP method, which exceeds the $95\%$ achieved by MOBO. Table \ref{tab:solution_comp} also shows that MORBO was particularly successful in finding solutions that exceed the value of lysine achieved by MFP ($16\%$ more for MORBO) and also that, in comparison to MOBO, a smaller percentage of solutions in MORBO do not improve any of the objectives. However, when it comes to the analysis of pairs of objectives, MOBO outperforms MORBO for CL and CE.

Regarding the quality of the solutions, Table \ref{tab:solution_distance} shows the average percentage improvement of MOBO and MORBO solutions compared to MFP \cite{pena2009multiobjective}, using the distance metric defined in Section \ref{sec:exper_setup}. The results are presented in percentage terms for the same objective configurations listed in Table \ref{tab:solution_distance}, and for 40 and 50 iterations. It is important to note that positive values indicate that MOBO (or MORBO, as applicable) solutions achieve, on average, an improvement over the MFP solution values, while negative values indicate that MFP surpassed MOBO or MORBO. Overall, from Table \ref{tab:solution_distance}, it can be observed that when MORBO finds solutions that outperform MFP in one or more objectives, the average percentage improvement is generally smaller than that of MOBO. Interestingly, when one of the objectives does not achieve any improvement, MORBO manages the losses better than MOBO. This behavior could be explained by MOBO's tendency to explore closer to the boundaries of the search space, potentially boosting some objectives significantly at the expense of others. Such behavior is less pronounced in the inner parts of the search space, which MORBO tends to explore more consistently.

\begin{table}[h]
\centering
\caption{Average percentage of improvement of MOBO and MORBO solutions in comparison to MFP \cite{pena2009multiobjective}}\label{tab:solution_distance}
\resizebox{\textwidth}{!}{\begin{tabular}{ccccccccccccc}
\toprule
& \multicolumn{6}{c}{MOBO} & \multicolumn{6}{c}{MORBO} \\
\cmidrule(rl){2-7} \cmidrule(rl){8-13} 
\multicolumn{1}{c}{$k$} & \multicolumn{3}{c}{\textbf{40}} & \multicolumn{3}{c}{\textbf{50}} & \multicolumn{3}{c}{\textbf{40}} & \multicolumn{3}{c}{\textbf{50}}  \\
\cmidrule(rl){1-1}  \cmidrule(rl){2-4} \cmidrule(rl){5-7} \cmidrule(rl){8-10} \cmidrule(rl){11-13} 
\textbf{Objective} & Cost ($\%$) &  Lys ($\%$) &  Ene ($\%$) &  Cost ($\%$) &  Lys ($\%$) &  Ene ($\%$) &  Cost ($\%$) &  Lys ($\%$) &  Ene ($\%$) &  Cost ($\%$) &  Lys ($\%$) &  Ene ($\%$)  \\
\midrule
C   &  14.07 &    0.00 &  -11.25 &  13.15 &    1.30 &  -12.61 &   7.89 &   -0.69 &   -9.54 &   8.18 &   -0.88 &   -9.27 \\
 L  & -13.81 &   31.73 &   -8.59 & -12.74 &   31.31 &  -10.44 & -20.25 &   27.46 &   -5.31 & -21.15 &   30.30 &   -4.87 \\
E   &   2.53 &    2.95 &   26.85 &   2.71 &    2.42 &   26.20 & -22.48 &   25.53 &    8.99 & -23.43 &   28.20 &    9.32 \\
CL  &   5.34 &   21.63 &  -38.77 &   4.88 &   22.23 &  -39.11 &   4.33 &    4.90 &   -9.45 &   4.34 &    5.01 &   -9.46 \\
CE  &  13.36 &  -10.19 &   18.96 &  12.85 &  -10.24 &   19.07 &   4.86 &   -1.65 &    7.53 &   5.58 &   -2.48 &    6.97 \\
 LE &  -8.57 &   19.45 &   34.50 &  -8.27 &   19.56 &   33.37 & -25.67 &   29.37 &    8.61 & -27.45 &   32.98 &    9.14 \\
CLE &   3.79 &   14.77 &   10.71 &   3.79 &   14.77 &   10.71 &   2.77 &    2.95 &    5.96 &   2.81 &    2.76 &    5.61 \\
\bottomrule
\end{tabular}}
\end{table}

\begin{figure}
    \centering
    \includegraphics[scale=0.3]{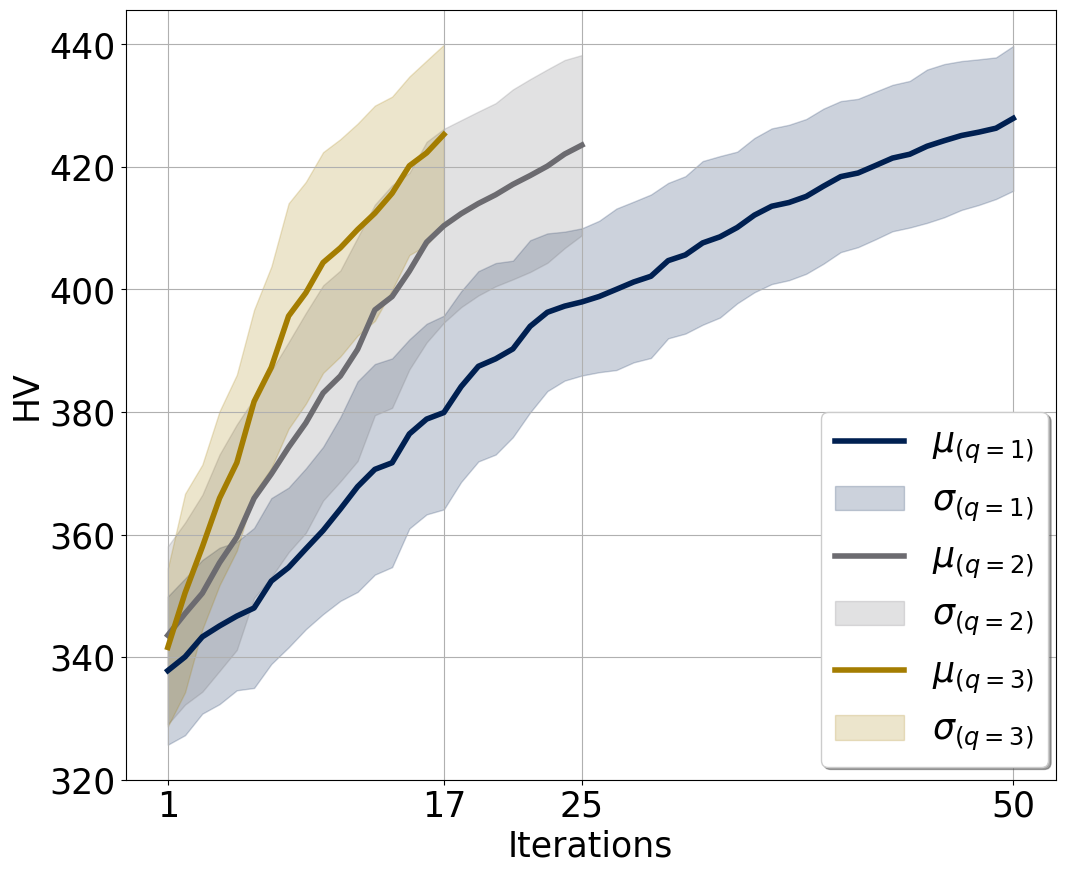}
    \caption{Means and standard deviations of HV values for $q=1,2,3$}\label{fig:batch}
\end{figure}

Lastly, Figure \ref{fig:batch} presents the HV values achieved by MORBO when generating batches of solutions during the optimization process. Unlike the results from MOBO in the context of swine diet design \cite{guribe2024}, MORBO shows increased convenience when $q>1$. The figure compares the evolution of HV under the condition of achieving a maximum of 50 query samples, which is the reference point for $q=1$. This condition is met in 25 and 17 iterations for $q=2$ and $q=3$, respectively. It can be observed that the improvement is progressive, and for $q=3$, MORBO reaches similar HV values in 17 iterations as those obtained for $q=1$ in 50 iterations. This result does not mean an improvement in the optimization budget or experimental cost associated with diet validation but a reduction in the time required to achieve a similar result.

Regarding the quality of the solutions obtained for  $q > 1$, the results confirm the success of batch querying. In 7 out of 30 runs, MORBO found solutions that dominated MFP within the first 10 iterations for both $q = 2$ and $q = 3$. This is just one run fewer than what was achieved for $ q = 1$ using the same set of seeds (see Table \ref{tab:solution_comp}). When considering all executions and reaching 50 new candidate solutions, MORBO found dominating solutions in 12 and 10 runs for $q = 2$ and $q = 3$, respectively. This represents a decrease of 29.41\% and 41.18\% compared to $q = 1$. However, given the iteration constraints of the diet design problem, the ability to speed up the optimization process for the first 20 new query solutions without significantly affecting solution quality is a noteworthy achievement.

\section{Conclusions}\label{sec:concl}

This work evaluated the use of a regionalized strategy to enhance the exploration and quality characteristics of the solutions provided by a multi-objective Bayesian optimization algorithm for diet design in the context of animal production. The setup consisted of a predefined three-objective optimization problem, including Lysine, Digestible Energy, and Cost, within a 17-dimensional input space representing ingredient proportions. This research advances the adaptation of BO methods to address food diet design problems, offering significant benefits for integrating multiple information sources that conventional metabolic-based methods cannot manage.

Results indicated that MORBO demonstrates a more efficient exploration of the search space, as evidenced by UMAP projections of the Pareto set, distances between consecutive iterative solutions, and diversity indicators, even though its HV values are lower than those of standard MOBO. The regionalized strategy proved four times more effective in finding solutions that dominate the MFP, a notable improvement given the critical optimization budget constraints in animal nutrition diet design. Moreover, the greater diversity of the Pareto front approximation facilitates better decision-making when selecting a single solution for practical implementation. In contrast, the non-dominated solutions provided by MOBO were more similar to each other, reducing the quality of the Pareto front approximation. MORBO's diversity is also reflected in the ingredient proportions of the best solutions compared to those found by MOBO (see Appendix A here and in \cite{guribe2024}).

Despite these advantages, MOBO solutions achieved a higher average percentage improvement over MFP, likely due to MOBO's tendency to explore the search space boundaries. However, MORBO was eight times more effective in finding non-dominated solutions compared to MFP during the first 10 iterations, a crucial factor for deploying a solution in the swine diet design context. Additionally, experiments with batches of query candidate solutions showed that MORBO could accelerate without compromising solution quality during the first 20 iterations.

Further experiments using data from pig farms should be carried out to validate the results and findings of this work. Additionally, exploring complementary strategies to improve the quality of the Pareto front approximation would be beneficial. This could include investigating alternative kernel functions for the GP covariance priors. The complexity of selecting an appropriate kernel for the surrogate GP increases in high-dimensional search spaces, as the objective function may become heterogeneous. For instance, employing non-stationary kernels that can model objective functions constant over large regions of the search space would be particularly useful \cite{eriksson2019scalable}. Implementing practical solutions for swine diet design also requires considering additional factors, such as incorporating contextual (non-controllable) variables \cite{krause2011contextual} and handling data from various farms and fidelity levels \cite{belakaria2021output}.

\section*{Acknowledgments}

G.D. Uribe-Guerra is supported by Colombia's Ministry of Science, Technology, and Innovation through Bicentennial Doctoral Excellence Scholarship Program - Court 2, 2020. J.D. Arias-Londoño started this work at the Antioquia University and finished it supported by a María Zambrano grant from the Universidad Politécnica de Madrid, Spain. 
The Autonomous Community of Madrid partially funded this work through the ELLIS Unit Madrid.
The authors would like to thank Mauricio Agudelo and Andrés Acevedo from \href{https://bialtec.co}{Bialtec} for their support in discussing the swine diet formulation requirements during the development of this work.

\section*{CRediT authorship contribution statement}

\textbf{Gabriel D. Uribe-Guerra:} Conceptualization, Software, Validation, Formal analysis, Data curation, Writing – original draft, Visualization. \textbf{Danny Múnera-Ramírez:} Conceptualization, Methodology, Formal analysis, Writing – review \& editing, Supervision. \textbf{Julián D. Arias-Londoño:} Conceptualization, Methodology, Formal analysis, Investigation, Writing – review \& editing, Supervision, Project administration, Funding acquisition.

\section*{Data availability}

The code necessary to reproduce the experimental findings can be found at \url{https://github.com/jdariasl/morbo}

\appendix
\section{Ingredients and nutrients}\label{sec:appe1}

Tables \ref{tab:ingredients_mobo} and \ref{tab:nutrients_mobo} list the distribution of ingredients and their equivalent nutritional content, respectively, corresponding to the best solutions found by MORBO and the reference solution MFP proposed in \cite{pena2009multiobjective}.

\begin{table}[H]
    \centering
    \caption{Corresponding values of ingredients (in \%) for the best MFP and MORBO solutions}
    \label{tab:ingredients_mobo}
    \resizebox{\textwidth}{!}{\begin{tabular}{l|c|c|c|c|c}
    \hline
        {} & {\bf MFP}& \multicolumn{4}{c}{{\bf MORBO}}\\ \hline
        {\bf Ingredients}&$\left[151.4, 1.02, 14.31\right]$ & $\left[149.2, 1.03, 14.45\right]$ & $\left[149.91, 1.02, 14.35\right]$ & $\left[150.23, 1.05, 14.35\right]$ & $\left[149.43, 1.03, 14.32\right]$\\ \hline

Barley        &  13.53  &  11.38 &  12.44 &  10.80 &  12.42 \\
Wheat         &  22.25  &    20.53 &  25.17 &  22.44 &  25.15 \\
Corn          &  0.00  &      2.12 &   1.43 &   7.36 &   5.05 \\
Alfalfa       &  0.00  &     0.40 &   0.33 &   0.23 &   0.67 \\
Cassava Meal  &  0.00  &     0.42 &   0.27 &   0.14 &   1.52 \\
Soybean meal  &  15.08  &    16.46 &  16.40 &  17.37 &  16.45 \\
Fish meal     &  0.00  &     0.08 &   0.41 &   0.06 &   0.42 \\
Gluten feed   &  3.74  &     7.37 &   6.27 &   4.31 &   4.99 \\
Calcium Carbonate & 0.94 &     3.49 &   3.99 &   3.25 &   3.57 \\
Lysine 78\%    & 0.00   &     0.03 &   0.12 &   0.12 &   0.12 \\
Sunflower meal & 0.00  &   0.74 &   2.40 &   2.72 &   1.83 \\
Animal fat     & 0.00  &     0.48 &   0.62 &   0.42 &   0.23 \\
Beet pulp      & 0.00  &    0.78 &   1.04 &   0.62 &   0.71 \\
Lupin          & 10.00  &   8.75 &   6.93 &   5.59 &   6.04 \\
Peas           & 14.06  &    10.27 &   4.32 &   7.74 &   5.76 \\
Rye            & 20.00  &    16.64 &  18.14 &  15.97 &  14.93 \\
Dicalcium      & 0.4  &    0.17 &   0.14 &   0.28 &   0.14 \\ \hline
\end{tabular}}

\end{table}

\begin{table}[H]
    \centering
    \caption{Nutrient content of the best MFP and MORBO solutions.}
    \label{tab:nutrients_mobo}
    \resizebox{0.6\textwidth}{!}{\begin{tabular}{l|c|c|c|c|c}
    \hline
        {} & {\bf MFP}& \multicolumn{3}{|c|}{{\bf MORBO}}\\ \hline
        
       { Cost (\euro/MT)}  & $151.4$&$149.2$&$149.91$&$150.23$& $149.43$\\
       { Lysine ($\%$)}&$1.02$&$1.03$&$1.02$&$1.05$& $1.03$\\
       { Energy (MJ/kg)}  &$14.31$&$14.45$&$14.35$&$14.35$ & $14.32$\\
       { Crude Fibre ($\%$)} & $5.09$ &   $5.32$ &   $5.21$ &   $5.02$ &   $5.06$ \\
       { Calcium ($\%$)} & $0.60$ &   $1.53$ &   $1.72$ &   $1.46$ &   $1.56$ \\
       { Dry matter ($\%$)} & $89.15$&   $88.95$ &  $89.35$ &  $88.27$ &  $89.14$ \\
       { Crude protein ($\%$)} &$19.33$&  $19.30$ &  $18.78$ &  $18.74$ &  $18.48$ \\
       { Phosphorus ($\%$)}  &$0.48$&$0.45$ &   $0.45$ &   $0.46$ &   $0.44$ \\
       { Methionine+cystine ($\%$)}  &$0.60$& $0.61$ &   $0.62$ &   $0.61$ &   $0.60$ \\
       { Tryptophan ($\%$)}  &$0.21$& $0.21$ &   $0.21$ &   $0.21$ &   $0.21$ \\
       { Threonine ($\%$)}  &$0.7$&$0.70$ &   $0.67$ &   $0.67$ &   $0.66$ \\
       { Available phosphorus ($\%$)}   &$0.16$&$0.16$&$0.17$&$0.15$ & $0.16$\\ \hline

    \end{tabular}}
    
\end{table}

\bibliographystyle{elsarticle-num} %  
\bibliography{main}

\end{document}